\pdfoutput=1

\documentclass[11pt]{article}

\usepackage[final]{acl}
\usepackage{times}
\usepackage{latexsym}

\usepackage{comment}
\usepackage{hyperref}
\usepackage{url}
\usepackage{graphicx} 
\usepackage{multirow}
\usepackage{ragged2e}
\usepackage{booktabs}
\usepackage{adjustbox}
\usepackage{todonotes}
\usepackage{subcaption}
\usepackage{amsmath}
\usepackage{xcolor}
\usepackage{adjustbox}

\usepackage[T1]{fontenc}

\usepackage[utf8]{inputenc}

\usepackage{microtype}

\usepackage{inconsolata}

\usepackage{graphicx}

\title{Multi-Novelty:Improve the Diversity and Novelty of Contents Generated by Large Language Models via inference-time Multi-Views Brainstorming}

\author{
    \textbf{Arash Lagzian\textsuperscript{1}}, 
    \textbf{Srinivas Anumasa\textsuperscript{1}}, 
    \textbf{Dianbo Liu\textsuperscript{1}} \\
    \textsuperscript{1}National University of Singapore \\
    \texttt{alagzian@visitor.nus.edu.sg} \\
    \texttt{\{srinu\_pd, dianbo\}@nus.edu.sg} \\
}
\begin{document}
\maketitle
\begin{abstract}
Large Language Models (LLMs) demonstrate remarkable proficiency in generating accurate and fluent text. However, they often struggle with diversity and novelty, leading to repetitive or overly deterministic responses. These limitations stem from constraints in training data, including gaps in specific knowledge domains, outdated information, and an over-reliance on textual sources. Such shortcomings reduce their effectiveness in tasks requiring creativity, multi-perspective reasoning, and exploratory thinking, such as LLM based AI scientist agents and creative artist agents . To address this challenge, we introduce inference-time  multi-view brainstorming method, a novel approach that enriches input prompts with diverse perspectives derived from both textual and visual sources, which we refere to as 
\textit{"Multi-Novelty"}. By incorporating additional contextual information as diverse starting point for chain of thoughts, this method enhances the variety and creativity of generated outputs. Importantly, our approach is model-agnostic, requiring no architectural modifications and being compatible with both open-source and proprietary LLMs.

\end{abstract}

\begin{figure*}[t]
    \centering
    \includegraphics[width=\textwidth]{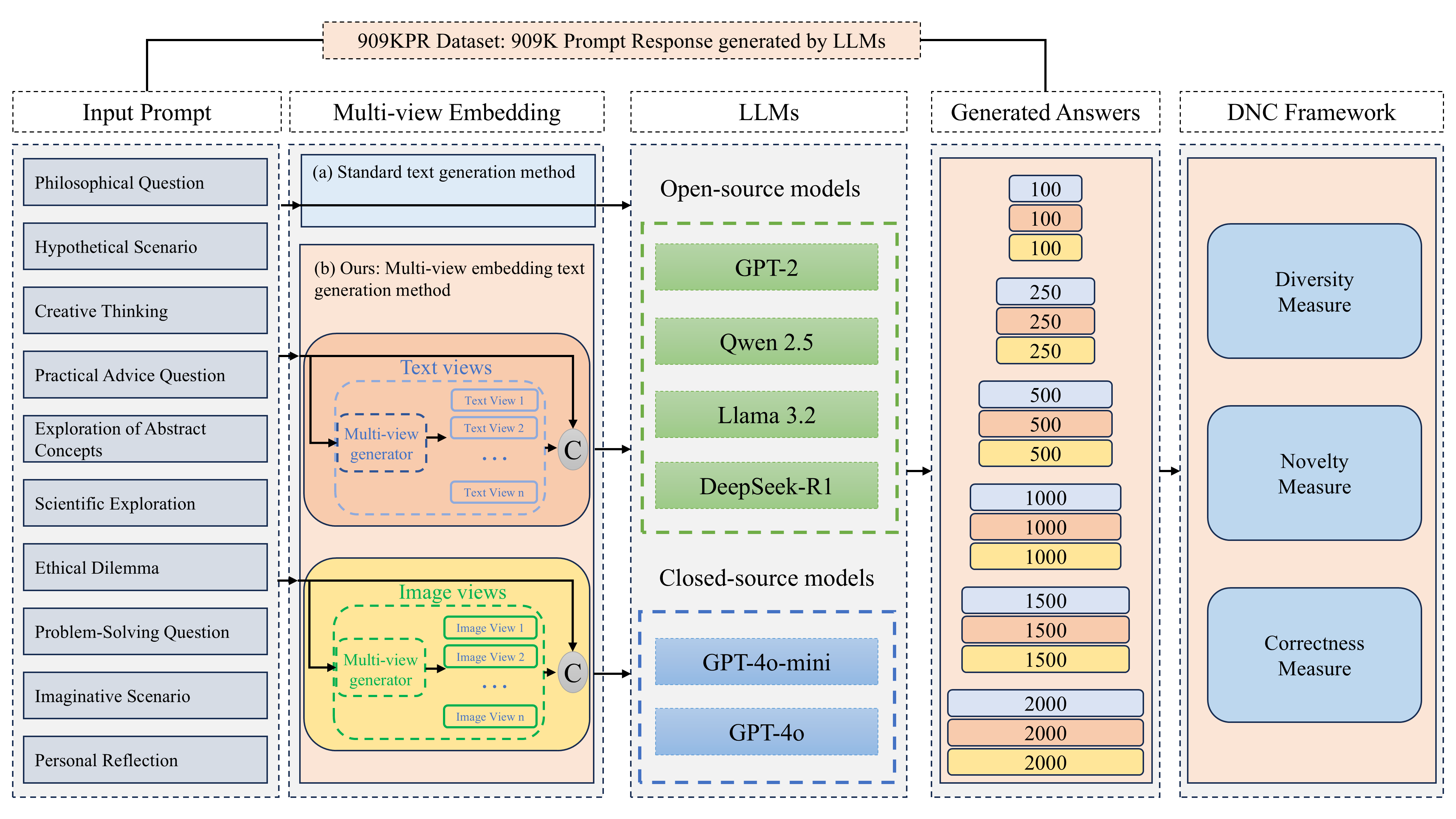} 
    \caption{Overview of our proposed methodology for enriching the diversity and novelty of LLM-generated content using multi-view embeddings. Our approach starts with diverse input prompts (e.g., philosophical questions, problem-solving scenarios, imaginative tasks) and generates multiple textual and visual views (n=50 per each prompt) through multi-view generators. These multi-view embeddings are then fed into various open-source and closed-source LLMs, such as Qwen, DeepSeek-R1, and GPT-4o, to generate responses with enriched diversity and novelty. We created the 909kPR dataset consisting of 909,500 generated answers across different models. Finally, the DNC framework evaluates the generated responses using three measures: diversity, novelty, and correctness.}
    \label{fig:overview}
\end{figure*}

\section{Introduction}

Rapid advances in large language models (LLMs) have spurred an ongoing debate on the usefulness of these models on tasks that require human-level creativity such as LLM based AI scientist agents and art or design assistant. On the one hand, LLMs have already been used in creative writing \cite{Bellemare2024, Orwig2024},
poetry generation \cite{porter2024}, idea generation \cite{lee2024, si2024can} and even originality \cite{Guzik2023}. On the other hand, some studies suggest that LLM creativity is significantly weaker than human creativity \cite{Chakrabarty2024} and that LLM-generated stories exhibit observable shortcomings \cite{sato2023, levenson2023}. A recent user study found that, the use of an AI assistant in writing appears to enhance the creativity of individual writers, however, it reduces the collective diversity of novel content produced by multiple writers \cite{Doshi2024}. Recent studies on AI scientists suggest novelty as a major issue in using LLM for scientific study \cite{lu2024ai}. This suggests that we should examine the distribution of
LLM creations for a given prompt instead of each
creation individually. Language models like GPT-4o \cite{openai2023gpt4} can produce repetitive answers in story telling and the answer not the same as original text \cite{xu2024echoes}. 

Existing methods primarily rely on input prompts to generate responses, often resulting in limited diversity. To address this, we propose leveraging multiple views for prompting to enhance response diversity. However, ensuring the correctness of these diverse responses is equally crucial. Recognizing the importance of both novelty and accuracy, we introduce a novel framework designed to generate diverse and novel responses while also providing a mechanism to evaluate these aspects. \\
Our key contributions in this work are as follows:
\begin{itemize}
\item We propose an architecture-independent approach to enrich generated text in terms of both novelty and diversity. By incorporating multiple views of the text embedding or image embedding , our method encourages the model to produce more diverse and novel outputs.
\item We introduce a framework to quantitatively assess the responses generated based on diversity, novelty, and correctness.
\item We conducted extensive experiments to demonstrate the effectiveness of our approach, evaluating over $909k$ generated responses and showcasing its improvements over existing LLM models including GPT-4o and DeepSeek-R1
\end{itemize}

\section{Related Work}
Creative writing \cite{Kobak2024, lee2024} is on the rise; however, some studies suggest that content generated by human users tends to be more creative \cite{Kefford2023}. This study show that ChatGPT's ideas are more purchased from Wharton MBA students. There is an ongoing debate about whether LLMs can enhance creativity. To explore this, \cite{lee2024} demonstrates that when participants were tasked with generating creative ideas for everyday purposes, their creativity improved. However, \cite{Begus2023} finds that AI-generated narratives often lack imagination and typically include plot twists in a more casual manner. Additionally, \cite{Chakrabarty2024} invited expert writers to evaluate stories generated by LLMs versus those created by professional writers using a standard creativity test. Their findings suggest that LLM-generated stories are less creative compared to those written by professionals.
Empirical studies have underscored this issue. For example, \cite{si2024can} conducted qualitative analyses involving human judgment and found that after generating 500 samples, 50\% were non-repetitive ideas. However, in the following 1,500 generations, only an additional 50\% of non-repetitive ideas were produced. Alarmingly, in the final 2,000 rounds, just 12.5\% of the generated ideas were non-repetitive. This suggests that while an individual LLM output may appear novel, when generating multiple outputs, the LLM tends to become repetitive, lacking the diversity necessary to effectively enhance collective creativity.
This decline underscores the resource inefficiency and diminishing returns in prolonged LLM-generated content. \\
\cite{McCoy2023} suggests that novelty in LLM outputs can be detected by ensuring "the text must not have been copied from the training data." However, a more recent study by \cite{xu2024echoes} argues that this definition is superficial. In their experiment on story continuation, they demonstrate that while GPT-4ogenerated samples may meet this standard, the generated continuations are still quite conventional and lack diversity.
\cite{Shaib2024} analyze different existing scores that can help measure diversity in LLM outputs, but these metrics all focus on surface-level features such as n-gram overlaps. \cite{Ghosal2022} indicate that "identifying novel text is not straightforward because the text many have less lexical overlap yet convey the same information." and to the best of our knowledge there is no study to evaluate the diversity, novelty, and correctness of the generated outputs at the same time.

\section{Proposed Method}

\subsection{Multi-view Embedding}
The inability of existing LLM models to generate diverse and novel text persists even after fine-tuning the temperature parameter. We propose that instead of solely adjusting the temperature, prompting models from multiple perspectives can effectively encourage the generation of more diverse and novel text. Figure \ref{fig:overview} offers an overview of our approach. Instead of directly interacting with the LLM model to generate a response, we first interact with a multi-view generator to create several perspectives of the given prompt. These generated views are then fed into the LLM model to produce the final response. The following section will give a more detailed overview of our approach.
In sections \ref{sec:text_view} and \ref{sec:image_view}, we explain the concepts of text view embedding and image view embedding within our method. Our experiments show that this approach enhances the model's understanding, increases output diversity, and boosts creativity.

\subsubsection{Text View Embedding}
\label{sec:text_view}

\begin{figure}[t]  
    \centering
    \includegraphics[width=\linewidth]{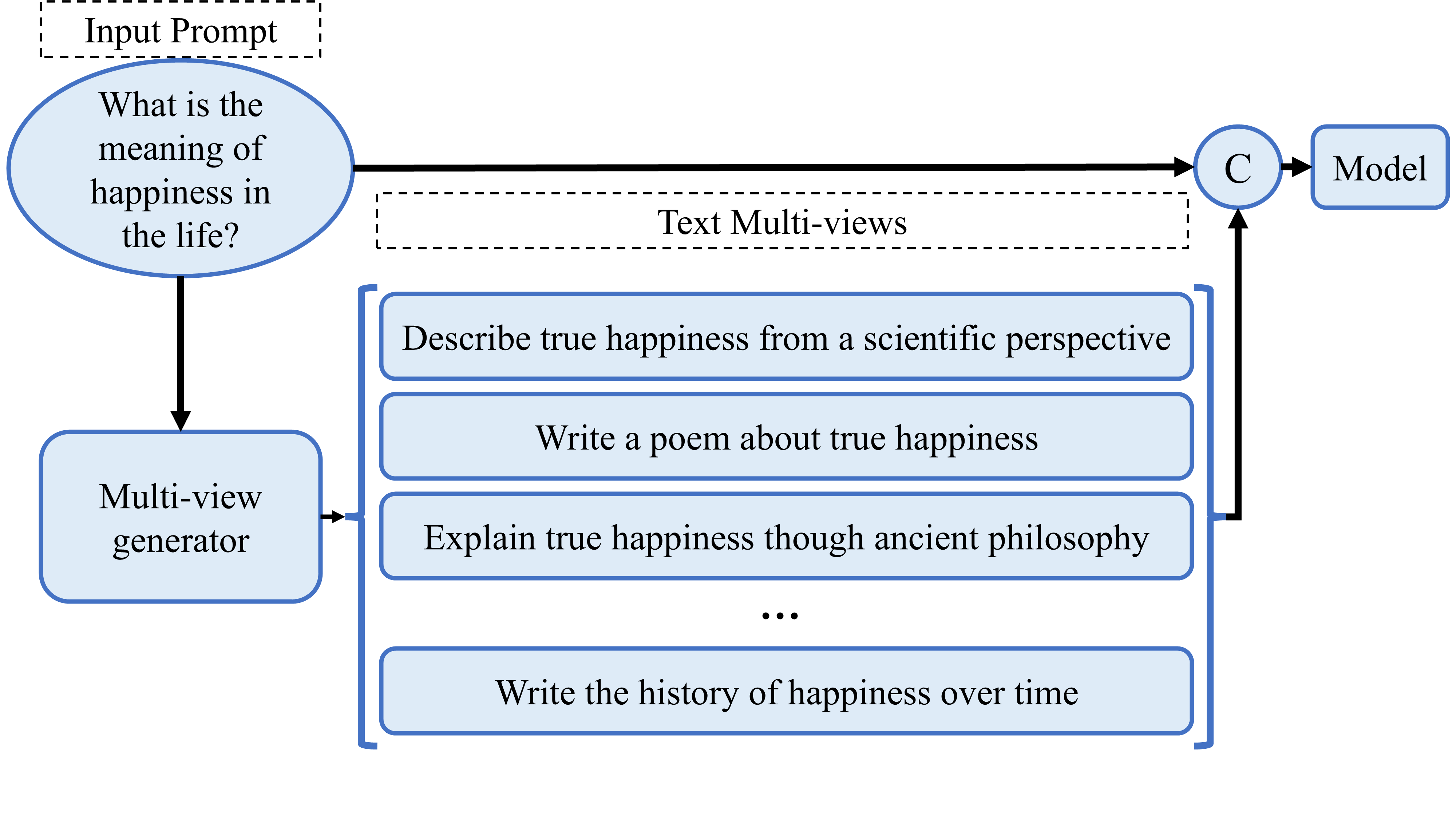}
    \caption{Text Multi-View Embedding.}
    \label{fig:text_views}
\end{figure}

Text Multi-view Embedding enhances the input prompt by generating multiple diverse perspectives or representations of the same concept, which are then combined before being fed into the model. This approach aims to provide a more comprehensive and context-rich input using various textual sources. These sources can be gathered from the internet, added manually, or even generated by a language model. In this work, we utilize GPT-4o as our text multi-view generator, with all texts in English. Figure \ref{fig:text_views} illustrates the text view embedding section of our method. In \cite{dipper2024}, GPT-4o was also used to extract diverse perspectives from math questions, resulting in improved performance on reasoning tasks.

\subsubsection{Image View Embedding}
\label{sec:image_view}
In addition to Text Multi-view Embedding, we also introduce Image Multi-view Embedding. Images contain rich contextual information and, unlike text, can offer diverse and multiple perspectives. 
Image Multi-view Embedding is used to enhance the input prompt by incorporating multiple image-based perspectives. This method starts by crawling for images related to the input prompt, which serve as visual representations of the concept. Once relevant images are retrieved, the Qwen-2VL \cite{qwen2vl} vision-language model is used to describe each image. These descriptions capture the visual content in textual form; however, they may lack consistency in writing style or contain structural issues. To improve the quality and coherence of these descriptions, we use a language model, in this case GPT-4o-mini, to rewrite and refine the original descriptions. The refined descriptions ensure that the textual representation of visual content is well-structured and stylistically consistent.  Instead of directly concatenating the descriptions with the input prompt, we use the refined descriptions as additional context when generating the final response. The model utilizes the detailed information from the rewritten image descriptions to provide richer, more accurate answers to the input prompt. The process of image view embedding is shown in the Figure \ref{fig:image_views} and Figure \ref{fig:happiness} in appendix \ref{appendix:a} shows an example of an image obtained by crawling over the internet using the input prompt, with its corresponding Image view and Answer.

\begin{figure}[ht!]
    \centering
    \includegraphics[width=0.48\textwidth]{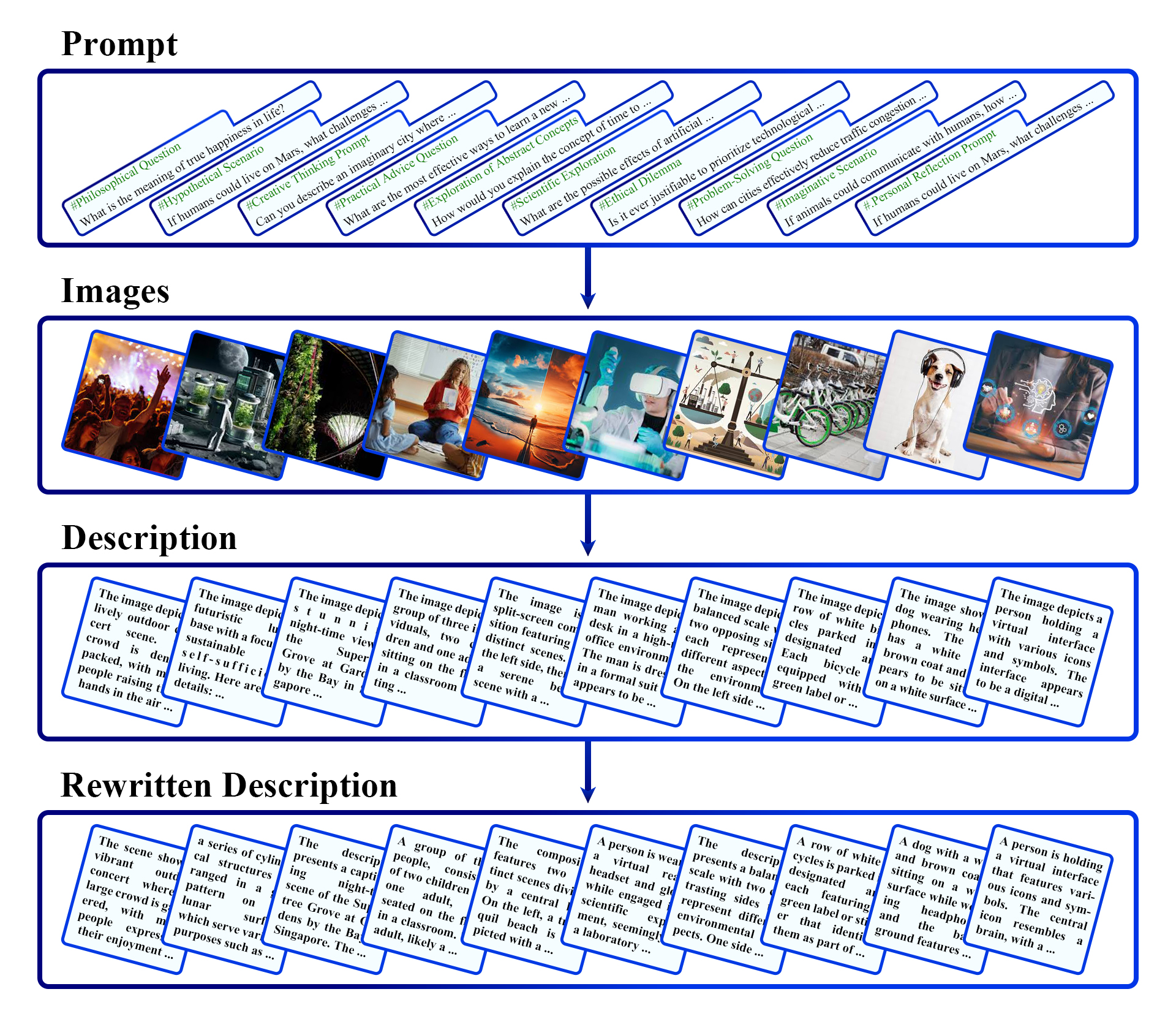} 
    \caption{This figure illustrates the process of preparing image view embeddings and provides an example for 10 input prompts. Row 1 displays 10 prompts from various subjects. Row 2 shows images crawled based on each input prompt. Row 3 presents the descriptions generated for the images, and Row 4 contains the rewritten descriptions, which serve as our image view embeddings.}
    \label{fig:image_views}
\end{figure}


\subsection{Metrics}
\label{sec:metrics}
To evaluate the responses generated by different LLMs for a given input prompt, it is crucial to consider multiple aspects of the generated text in order to quantify the model's performance. Existing works typically focus on one aspect—such as novelty, diversity, or correctness—individually. In contrast, our work takes a comprehensive approach by considering all these aspects to assess the model's performance.
\subsubsection{Diversity Measure}
\label{subsec:diversity_measure}
The diversity in the text generated by LLMs can be measured from various aspects. Two of the most important aspects are:

\begin{itemize} 
\item Diversity across different text responses. 
\item Diversity within different tokens generated in a single text response. 
\end{itemize}

In this work, first we try to describe each metric and then plot the importance of each one in diversity measurement and special cases for each metric and finally talk about the results.

\paragraph{MTLD:} Lexical diversity is a key measure of linguistic richness in generated text. To evaluate this, we use the Measure of Textual Lexical Diversity (MTLD) \cite{McCarthy2010}. MTLD calculates the mean length of text segments that maintain a predefined type-token ratio (TTR). This approach overcomes limitations of traditional TTR metrics, which are sensitive to text length. Following \cite{McCarthy2010}, we set the TTR threshold to 0.72, as this value has been empirically validated to balance sensitivity and robustness across a variety of text datasets.


\paragraph{Semantic Diversity of Text (SDT)}
We measure semantic diversity using Term Frequency–Inverse Document Frequency (TF-IDF) representations of all responses. First, each response is transformed into a TF-IDF vector based on term usage and distinctiveness. Then, we compute the pairwise cosine similarity among these TF-IDF vectors to quantify how similar the responses are. Because higher similarity implies lower diversity, we define our overall semantic diversity metric as:

\[
\text{SDT} = 1 - \overline{\text{Sim}},
\]

where \( \overline{\text{Sim}} \) is the average of all pairwise cosine similarities between the TF-IDF vectors of the responses. Consequently, larger values of SDT indicate greater variation in term usage—i.e., more semantic diversity. This formulation ensures that near-identical responses (high cosine similarity) yield low diversity scores, while dissimilar responses produce high diversity scores.

\paragraph{Semantic Diversity of Embeddings (SDE)}
BERT-based semantic diversity measures the variation in the \emph{semantic content} of generated text. 
Unlike lexical diversity, which focuses on surface-level differences in word usage, BERT embeddings capture 
deeper contextual differences by mapping sentences into a dense semantic space. We compute pairwise cosine 
similarity among the embeddings of all responses and then take the average cosine distance 
(i.e., \(1 - \text{cosine similarity}\)) to quantify diversity. Higher semantic diversity indicates a broader 
range of meanings and greater contextual richness. This approach leverages the contextualized representations 
of BERT~\citep{bert2019}, offering a robust and nuanced assessment of the diversity and coherence of 
model-generated outputs.
The \textbf{Semantic Diversity of Embeddings} (\(\textbf{SDE}\)) can be quantified by averaging the 
pairwise \emph{cosine distance} between all embeddings. Given a set of embeddings 
\(\{\mathbf{e}_1, \mathbf{e}_2, \dots, \mathbf{e}_n\}\), the cosine similarity between two embeddings 
\(\mathbf{e}_i\) and \(\mathbf{e}_j\) is defined as:
$
\text{CosineSimilarity}(\mathbf{e}_i, \mathbf{e}_j) 
= \frac{\mathbf{e}_i \cdot \mathbf{e}_j}{\|\mathbf{e}_i\| \|\mathbf{e}_j\|}.
$
The corresponding cosine distance is:
$
1 - \text{CosineSimilarity}(\mathbf{e}_i, \mathbf{e}_j).
$
Thus, the overall \textbf{SDE} is:
\[
\textbf{SDE} = \frac{2}{n(n-1)} \sum_{1 \leq i < j \leq n} 
\biggl(1 - \frac{\mathbf{e}_i \cdot \mathbf{e}_j}{\|\mathbf{e}_i\| \|\mathbf{e}_j\|}\biggr),
\]
where \(n\) is the total number of embeddings (responses).

\paragraph{Self-BLUE}
Self-BLEU evaluates the diversity of generated text by measuring how similar each generated sample is to the others. Unlike traditional BLEU, which compares a generated sample to reference texts, Self-BLEU treats each sample as a reference for the others. A high Self-BLEU score signals redundancy, reflecting repeated patterns or limited variability \cite{selfblue2018}. In our setting we subtract the mean value across all answers from 1 to measure the diversity across all answers.

\paragraph{Lexical Entropy}


It quantifies the uncertainty or variability in word choice within generated text, serving as a measure of lexical diversity. Higher entropy indicates a broader and more varied vocabulary, while lower entropy suggests repetitive or predicTable word usage. We calculate Shannon entropy (in bits) over all tokens in the generated responses, based on their relative frequencies. For a given set of responses, the probability of each unique token is determined, and the entropy is computed as: $
H = -\sum_{i} p_i \log_2 p_i
$ where \( p_i \) represents the probability of the \( i \)-th token. A higher lexical entropy signifies reduced repetition and a richer vocabulary, making it a valuable indicator of linguistic creativity in language models.


\subsubsection{Novelty Measure}
Novelty detection is critical for assessing large language model (LLM) outputs, as it differentiates existing knowledge from newly introduced information. This task can be framed as Natural Language Inference (NLI), wherein one determines whether a hypothesis contains content not found in a premise set. Prior studies have employed both NLI-based methods and embedding techniques such as SBERT \cite{Ghosal2022} to measure semantic similarity, illustrating the importance of novelty detection for refining LLM-generated text.

In our work, we wanted to assess the inherent capability of LLMs to perform as NLI models in detecting novel versus non-novel documents. To achieve this, we used \textbf{TAP-DNLD 1.0} \cite{Ghosal18} a dataset consisting of around $2.8k$ novel and $2.7k$ non-novel documents, and each document belongs to one of the ten categories and each target document labeld versus three source document by human. For comparison , first we sampled from this dataset in three different seeds and in each seed from each category sampled 5 novel and 5 non-novel document to have a balance data and totally 300 document then We used GPT-4o and SBERT as novelty detector to have better comparison and results reported in Table \ref{tab:classification_metrics_sbert_gpt}.

\begin{table}[ht]
    \centering
    \resizebox{\linewidth}{!}{ 
    \begin{tabular}{ l c c c c c }
        \hline
        \textbf{Seed} & \textbf{Model} & \textbf{Accuracy} & \textbf{Precision} & \textbf{Recall} & \textbf{F1-score} \\
        \hline
        \multirow{2}{*}{Seed0} 
        & SBERT  & 0.6700 & 0.6269 & 0.8400 & 0.7179 \\ 
        & GPT-4O & \textbf{0.6900} & \textbf{0.6415} & \textbf{0.8600}  & \textbf{0.7347 } \\ 
        \hline
        \multirow{2}{*}{Seed1} 
        & SBERT  & \textbf{0.7200} & \textbf{0.6528} & \textbf{0.9400} & \textbf{0.7705} \\ 
        & GPT-4O & 0.6600  & 0.6086  & 0.9000  & 0.7260  \\ 
        \hline
        \multirow{2}{*}{Seed2} 
        & SBERT  & 0.6900 & 0.6338 & 0.9000 & 0.7438 \\ 
        & GPT-4O & \textbf{0.6950 } & \textbf{0.6368 } & \textbf{0.9100}  & \textbf{0.7492 } \\ 
        \hline
    \end{tabular}
    }
    \caption{Classification metrics to evaluate novelty detectors GPT-4o and SBERT models on dataset \textbf{TAP-DNLD 1.0} \cite{Ghosal18}.}
    \label{tab:classification_metrics_sbert_gpt}
\end{table}

\begin{figure}[ht!]
    \centering
    \includegraphics[width=\linewidth]{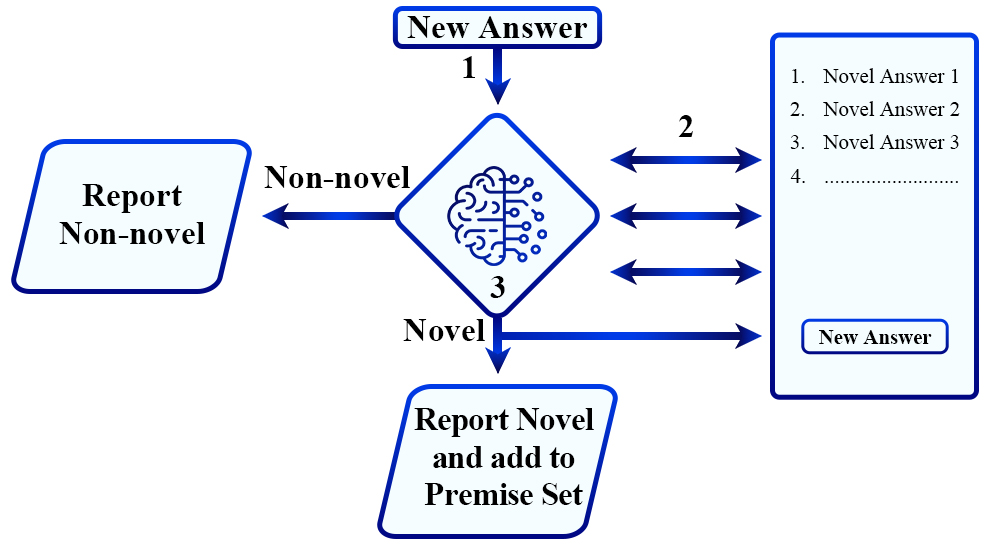} 
    \caption{Novelty detection method by using GPT-4o and SBERT.}
    \label{fig:novelty detector}
\end{figure}

After establishing the baseline performance of these models, we utilized GPT-4o and SBERT as novelty detectors to evaluate the novelty of generated outputs across different language models, as well as to assess the impact of our method on the novelty of generated responses. The novelty detection process is illustrated in Figure \ref{fig:novelty detector}. In this approach, for each set of prompt outputs, the first generated response is considered novel by default. Subsequent responses are sequentially analyzed by the novelty detector, where each new answer is compared against previously identified novel responses. This comparison is conducted in a hypothesis-premise framework, where the new response serves as the hypothesis, and previously identified novel answers form the premise set. If the new answer introduces additional information not present in the premise set, it is classified as a novel answer and added to the premise set. Otherwise, it is labeled as redundant.

\subsubsection{Correctness Measure}
\label{subsubsec: correctness_measure}
Answers could be novel or even diverse but completely irrelevant to the input prompt (see example in appendix \ref{sec:correctness_measure_example}), or they can have different structures. In this section we want to answer these two questions:
\begin{itemize}
    \item Is the generated answer correct and relevant to the given input prompt?  
    \item How well does the generated answer adhere to proper language structure and grammatical accuracy?  
\end{itemize}
To address the first question, we designed an experiment to assess whether language models can accurately detect relevant answers to a given prompt. For this evaluation, we used the GPT-WritingPrompts dataset \cite{gptstpries}, which contains approximately $97k$ unique prompts along with responses from both humans and GPT-3.5. However, in our study, we exclusively utilized the human-generated answers. 
We sampled data using ten different random seeds, selecting 1K samples per seed. These samples were then clustered into ten distinct groups. From each cluster, we randomly picked $100$ prompts. For each selected prompt, we assigned one relevant answer from the same cluster, labeling it as "Correct," and one irrelevant answer from a different cluster, labeling it as "Incorrect," ensuring that the chosen answers were distinct. Since the answers are approximately $500$ words long, we assume that if an answer is related to the prompt, its summary will also be relevant, and similarly, if an answer is irrelevant, its summary will remain unrelated. Based on this assumption, we use the GPT-4o model to generate summaries for both correct and incorrect answers, setting the maximum length to 250 words. Next, we evaluate the ability of two well-known LLMs, GPT-4o and DeepSeekV3 \cite{deepseekv3}, to determine the correctness of the answers. The results, presented in Table \ref{tab:mean_std_metrics_sum}, indicate that GPT-4o outperforms DeepSeekV3 in detecting correctness. An example prompt and its related responses are provided in Appendix \ref{appendix:b}. Based on these findings, we use GPT-4o for correctness detection in our experiments.
\begin{table}[ht]
    \centering
    \resizebox{\linewidth}{!}{ 
    \begin{tabular}{ c c c c c }
        \hline
        \textbf{Model} & \textbf{Accuracy} & \textbf{Precision} & \textbf{Recall} & \textbf{F1-score} \\
        \hline
        DeepSeekV3  & 0.8410 ± 0.0223 & \textbf{0.9564 ± 0.0285} & 0.7270 ± 0.0447 & 0.8191 ± 0.0301 \\
        \hline
        GPT-4O       & \textbf{0.8980 ± 0.0214} & 0.9468 ± 0.0255 & \textbf{0.8310 ± 0.0316} & \textbf{0.8946 ± 0.0228} \\
        \hline
    \end{tabular}
    } 
    \caption{Results on classification task to detect correct answer across all $10$ seeds for DeepSeekV3 and GPT-4o models. Complete results can be found in Table \ref{tab:classification_metrics_correctness}, Appendix \ref{appendix:a}.}
    \label{tab:mean_std_metrics_sum}
\end{table}


To address the second question, we designed an experiment to evaluate language models based on their ability to assign scores to generated English texts. For this evaluation, we utilized the IELTS Writing Task 2 dataset with labeled scores \cite{Mazlumi2023} from Kaggle. This dataset contains 642 responses for Task 1 and 793 responses for Task 2. Since only Task 2 is relevant to our study, we used its questions and answers to assess language models. We randomly sampled 100 responses in three different seeds, conducting two iterations to evaluate text quality. Two well-known models assigned scores between 1 and 9 (based on the IELTS writing score range), and the results are presented in Table \ref{tab:structure_score_comperession}. Based on this experiment, we selected the DeepSeekV3 model to evaluate responses in terms of grammar and overall English structure, assigning each answer a score between 1 and 10. The prompt templates for both experiments are provided in Appendix~\ref{appendix:b}. Based on this experiment we used DeepSeekV3 model to evaluate the outputs.
\begin{table}[ht]
    \centering
    \resizebox{\linewidth}{!}{ 
    \begin{tabular}{ c c c }
        \hline
        \textbf{Seed} & \textbf{GPT-4o} & \textbf{DeepSeekV3} \\
        \hline
        Seed0  & 2.3775 ± 0.0350 & \textbf{2.0775 ± 0.0025} \\
        \hline
        Seed1  & 2.2288 ± 0.0512 & \textbf{2.0300 ± 0.0050} \\
        \hline
        Seed2  & 2.1562 ± 0.0587 & \textbf{1.9050 ± 0.0500} \\
        \hline
    \end{tabular}
    } 
    \caption{MSE (Mean ± Std) for GPT-4o and DeepSeekV3 per seed for two iterations. Lower values are bolded.}
    \label{tab:structure_score_comperession}
\end{table}

\section{Experimental Results}
In Section \ref{subsec:diversity_measure}, we introduced various diversity metrics. In this section, we provide experimental results to assess the diversity of different language models. Additionally, we applied the multi-view embedding method to four open-source models—GPT-2 Medium \cite{gpt2}, Qwen2.5-1.5B \cite{qwen2.5-1.5b}, Llama 3.2-3B \cite{llama32} ,and DeepSeek-R1-7B \cite{deepseekr1-7b}—as well as two API-based models, GPT-4o \cite{openai2023gpt4} and GPT-4o Mini \cite{openai2024gpt4omini}\footnote{We conducted all expermiments on one A100 GPU and spent 750\$ for API models.}. Our experiments utilized 10 diverse prompts spanning multiple domains (provided in Appendix \ref{appendix:b}), instructing each model to generate responses. This process was repeated across different sample sizes ranging from $100$ to $2000$, with a fixed maximum sequence length of $125$, resulting in a total of $909500$ generated responses and for having fair compression we used the same parameters for all models like $tempereture = 0.9$ and $top\_k = 0.95$. The generated outputs were then evaluated in terms of diversity, novelty, and correctness. Table \ref{tab:mean_results_diversity} presents the diversity measurement results. The experimental findings demonstrate that our method enhances diversity across all models. In some cases, we observed up to a threefold increase in the diversity of generated outputs.

\begin{table}[ht]
    \centering
    \resizebox{\linewidth}{!}{
    \begin{tabular}{ l c c c c c }
        \hline
        \textbf{Models} & \textbf{MTLD} & \textbf{TF-IDF} & \textbf{Lexical Entropy} & \textbf{Semantic Embedding} & \textbf{Self-BLEU} \\
        \hline
        GPT-2                      & \textbf{41.36 ± 0.04}  & 0.87 ± 0.01  & 8.16 ± 0.09  & 0.55 ± 0.00  & 0.59 ± 0.11  \\
        GPT-2 + Text View          & 34.95 ± 1.14  & \textbf{0.90 ± 0.01}  & \textbf{8.35 ± 0.13}  & \textbf{0.65 ± 0.01}  & \textbf{0.65 ± 0.09}  \\
        \hline
        Qwen-2.5                       & 67.19 ± 0.14  & 0.80 ± 0.01  & 7.91 ± 0.05  & 0.22 ± 0.00  & 0.25 ± 0.09  \\
        Qwen-2.5 + Text View           & 67.64 ± 0.10  & \textbf{0.86 ± 0.01}  & \textbf{8.46 ± 0.07}  & \textbf{0.38 ± 0.10}  & \textbf{0.39 ± 0.10}  \\
        Qwen-2.5 + Image View          & \textbf{70.28 ± 0.04}  & 0.84 ± 0.01  & 8.30 ± 0.07  & 0.30 ± 0.00  & 0.37 ± 0.11  \\
        \hline 
        LLama-3.2                  & 52.45 ± 0.18 & 0.72 ± 0.01 & 7.41 ± 0.05 & 0.23 ± 0.00 & 0.18 ± 0.07\\
        LLama-3.2 + Text View      & 54.23 ± 0.05 & \textbf{0.84 ± 0.01} & \textbf{8.21 ± 0.09} & \textbf{0.43 ± 0.00} & \textbf{0.36 ± 0.11}\\
        LLama-3.2 + Image View     & \textbf{54.46} ± 0.07 & 0.76 ± 0.01 & 7.70 ± 0.06 & 0.31 ± 0.00 & 0.27 ± 0.07\\
        \hline
        DeepSeek-R1                & 52.51 ± 0.10  & 0.78 ± 0.01  & 7.64 ± 0.05  & 0.24 ± 0.00  & 0.24 ± 0.05  \\
        DeepSeek-R1 + Text View    & \textbf{54.61 ± 0.21}  & \textbf{0.85 ± 0.01}  & \textbf{8.21 ± 0.08}  & \textbf{0.39 ± 0.00}  & \textbf{0.36 ± 0.09}  \\
        DeepSeek-R1 + Image View   & 54.36 ± 0.05  & 0.83 ± 0.01  & 8.07 ± 0.06  & \textbf{0.39 ± 0.00}  & 0.34 ± 0.10  \\
        \hline
        GPT-4O Mini                & 57.95 ± 0.02  & 0.63 ± 0.01  & 7.06 ± 0.02  & 0.11 ± 0.00  & 0.09 ± 0.05  \\
        GPT-4O Mini + Text View    & \textbf{59.59 ± 0.11}  & \textbf{0.81 ± 0.03}  & \textbf{7.93 ± 0.03}  & \textbf{0.33 ± 0.00}  & \textbf{0.24 ± 0.11}  \\
        GPT-4O Mini + Image View   & 57.74 ± 2.59  & 0.79 ± 0.02  & 7.79 ± 0.03  & 0.30 ± 0.01  & 0.23 ± 0.08  \\
        \hline
        GPT-4O                     & 57.18 ± 0.71  & 0.66 ± 0.01  & 7.01 ± 0.14  & 0.11 ± 0.02  & 0.08 ± 0.04  \\
        GPT-4O + Text View         & \textbf{58.25 ± 0.03}  & 0.79 ± 0.01  & \textbf{7.86 ± 0.05}  & \textbf{0.31 ± 0.00}  & 0.23 ± 0.08  \\
        GPT-4O + Image View        & 55.75 ± 1.49  & \textbf{0.79 ± 0.01}  & 7.80 ± 0.08  & \textbf{0.32 ± 0.01}  & \textbf{0.26 ± 0.10}  \\
        \hline
    \end{tabular}
    }
    \caption{Mean ± standard deviation of diversity metrics across six different sample sizes ranging from $100$ to $2000$ per each prompt (10 prompts), with $max\_length=125$. Detailed results are provided in Appendix \ref{appendix:a}.}
    \label{tab:mean_results_diversity}
\end{table}

Table \ref{tab:novelty_gpt_ds} illustrates the novelty score of two well-known large language models GPT-4o and DeepSeek-R1 and impact of our method. By incorporating multi-view embeddings, we enriched the input representation, leading to the generation of more novel and diverse responses from these models. This analysis provides valuable insights into how multi-view embedding strategies influence novelty detection and enhance the creativity of LLM outputs. The results show that for GPT-4o, our approach in some cases led to around ninefold improvement in novelty when text or image view embeddings were applied. Similarly, for DeepSeek-R1—one of the most inherently creative models—our method resulted in approximately a twofold increase in the novelty score. Additional results for other models, evaluated solely using SBERT as the novelty detector, are presented in Table \ref{tab:novelty_sbert_2} in Appendix \ref{appendix:a}. 


\begin{table}
\centering
\resizebox{\linewidth}{!}{
\begin{tabular}{lcccccc}
\toprule
 & \multicolumn{2}{c}{num\_samples = 100} & \multicolumn{2}{c}{num\_samples = 250} & \multicolumn{1}{c}{num\_samples = 500} \\
\cmidrule(lr){2-3} \cmidrule(lr){4-5} \cmidrule(lr){6-6}
Model & GPT-4o & SBERT & GPT-4o & SBERT & SBERT \\
\midrule
GPT-4o & 10.60 & 5.4 & 7.28 & 3.52 & 2.9 \\
GPT-4o + Text View & \underline{29.30} & \underline{40.2} & \underline{20.4} & \textbf{32.08} & \textbf{24.48} \\
GPT-4o + Image View & \textbf{42.60} & \textbf{47.3} & \textbf{25.52} & \underline{30.44} & \underline{22.54} \\ \hline
DeepSeek-R1 & 27.40 & 43.6 & 18.6 & 36.24 & 31.82 \\
DeepSeek-R1 + Text View & \textbf{46.20} & \textbf{75.3} & \textbf{38.22} & \underline{63.88} & \underline{55.86} \\
DeepSeek-R1 + Image View & \underline{40.60} & \underline{73.7} & \underline{27.56} & \textbf{65.04} & \textbf{56.18} \\
\bottomrule
\end{tabular}}
\caption{Results on percentage of novelty score across different models according to two novelty detectors, GPT-4o and SBERT. More results for other models exist in Table \ref{tab:novelty_sbert_2}.}
\label{tab:novelty_gpt_ds}
\end{table}

\begin{figure*}[t]
  \centering
  \includegraphics[width=0.45\linewidth]{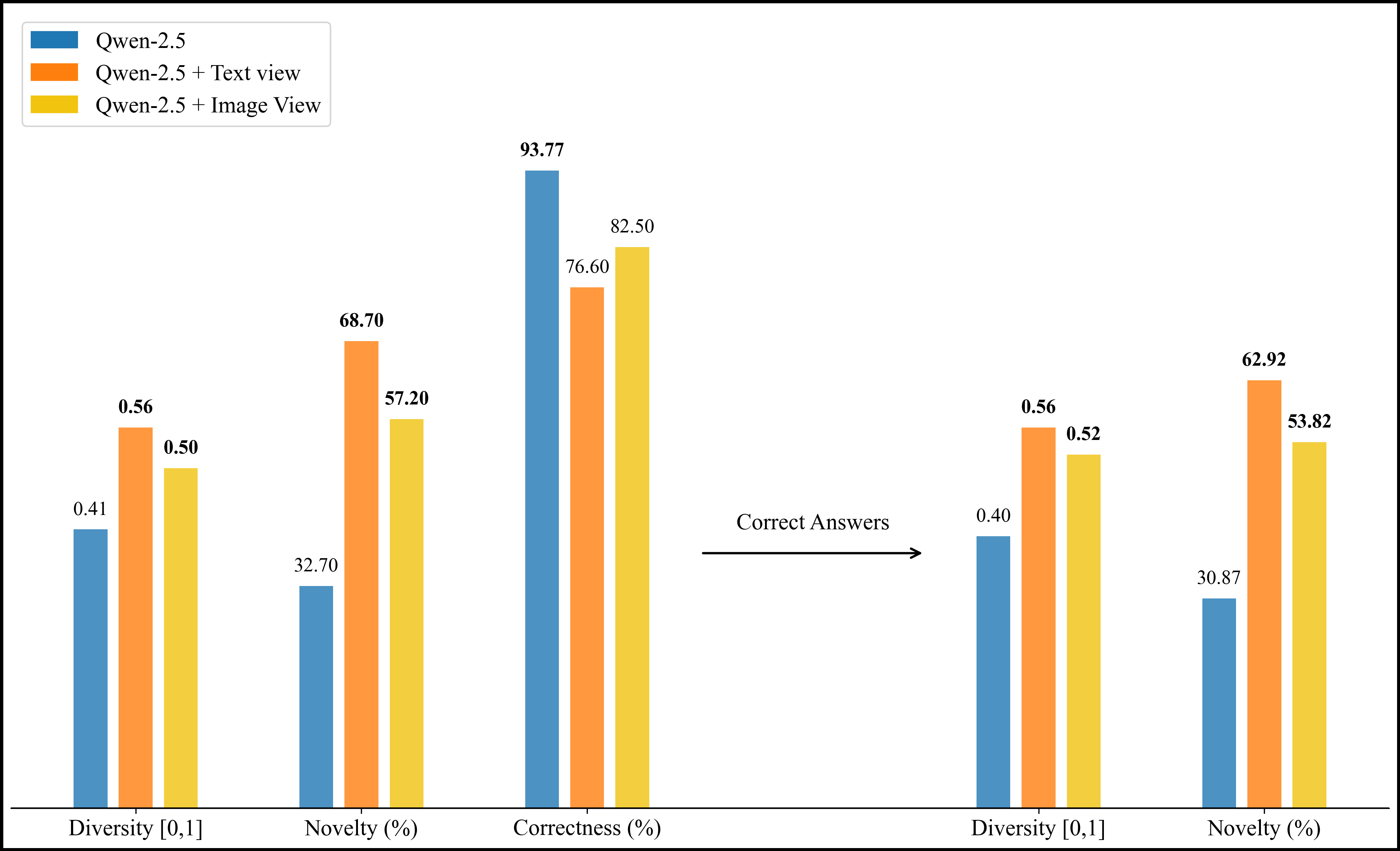} \hspace{3em}
  \includegraphics[width=0.45\linewidth]{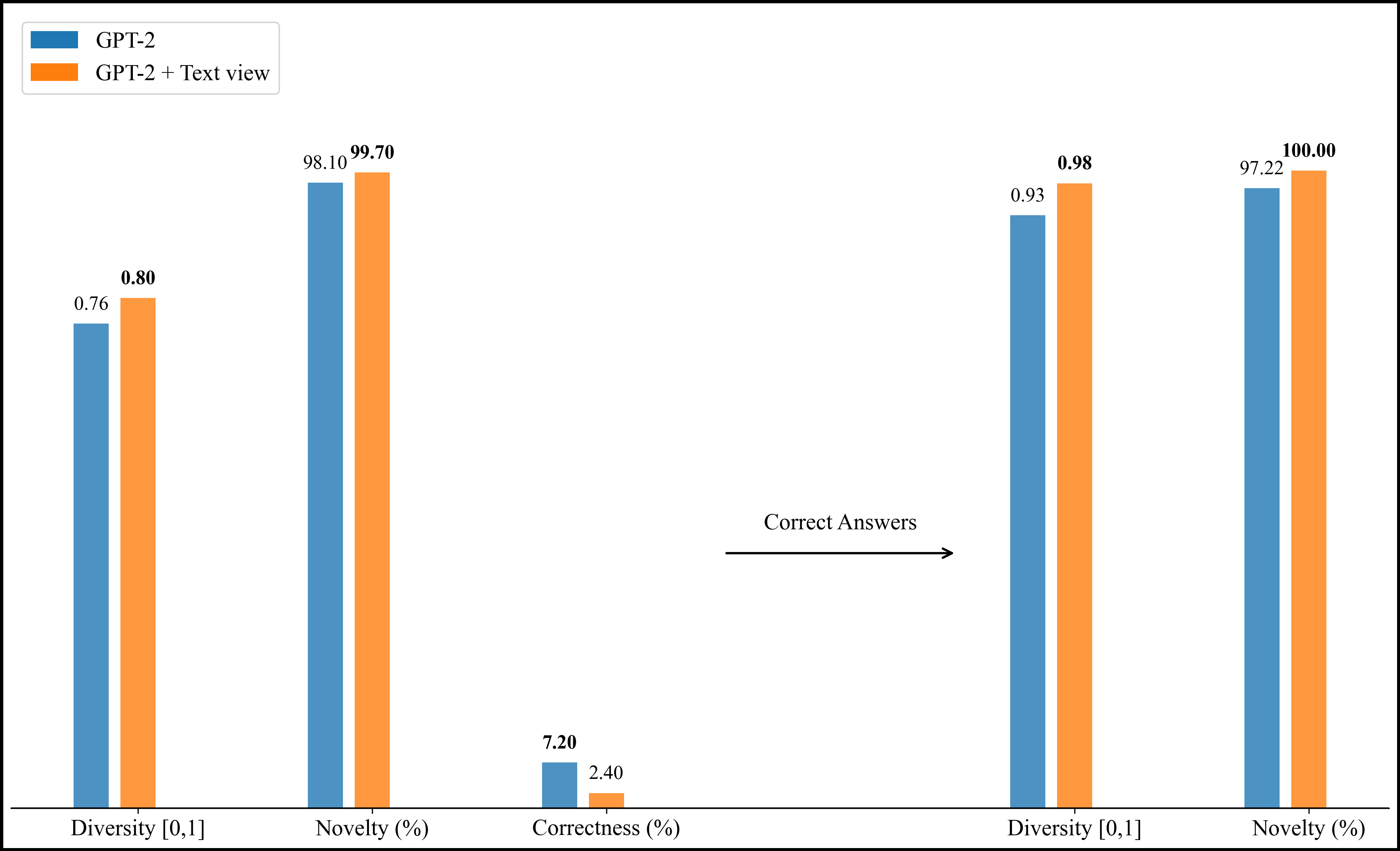} \\[+1mm]
  \includegraphics[width=0.45\linewidth]{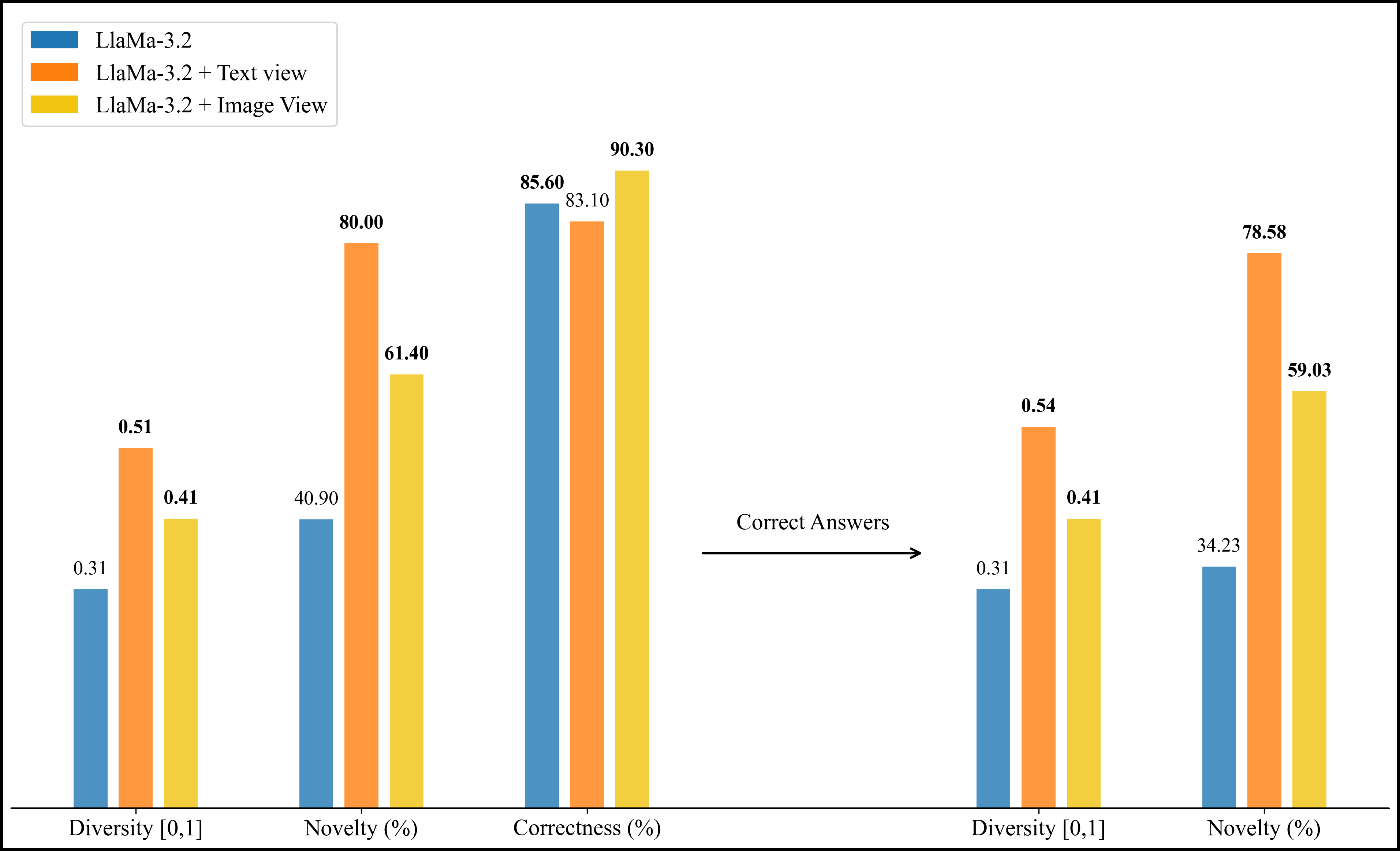}\hspace{3em}
  \includegraphics[width=0.45\linewidth]{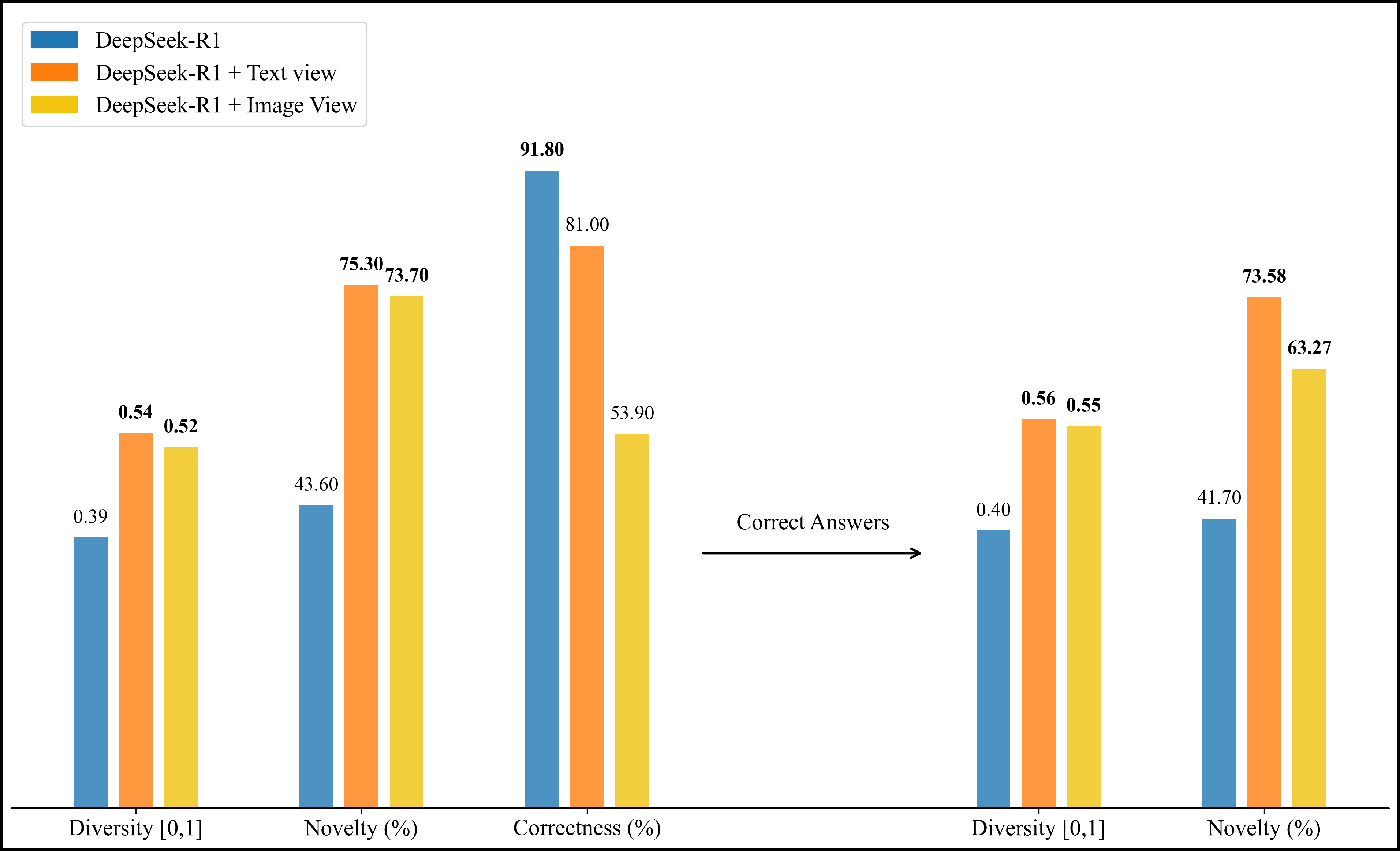} \\[+1mm]
  \includegraphics[width=0.45\linewidth]{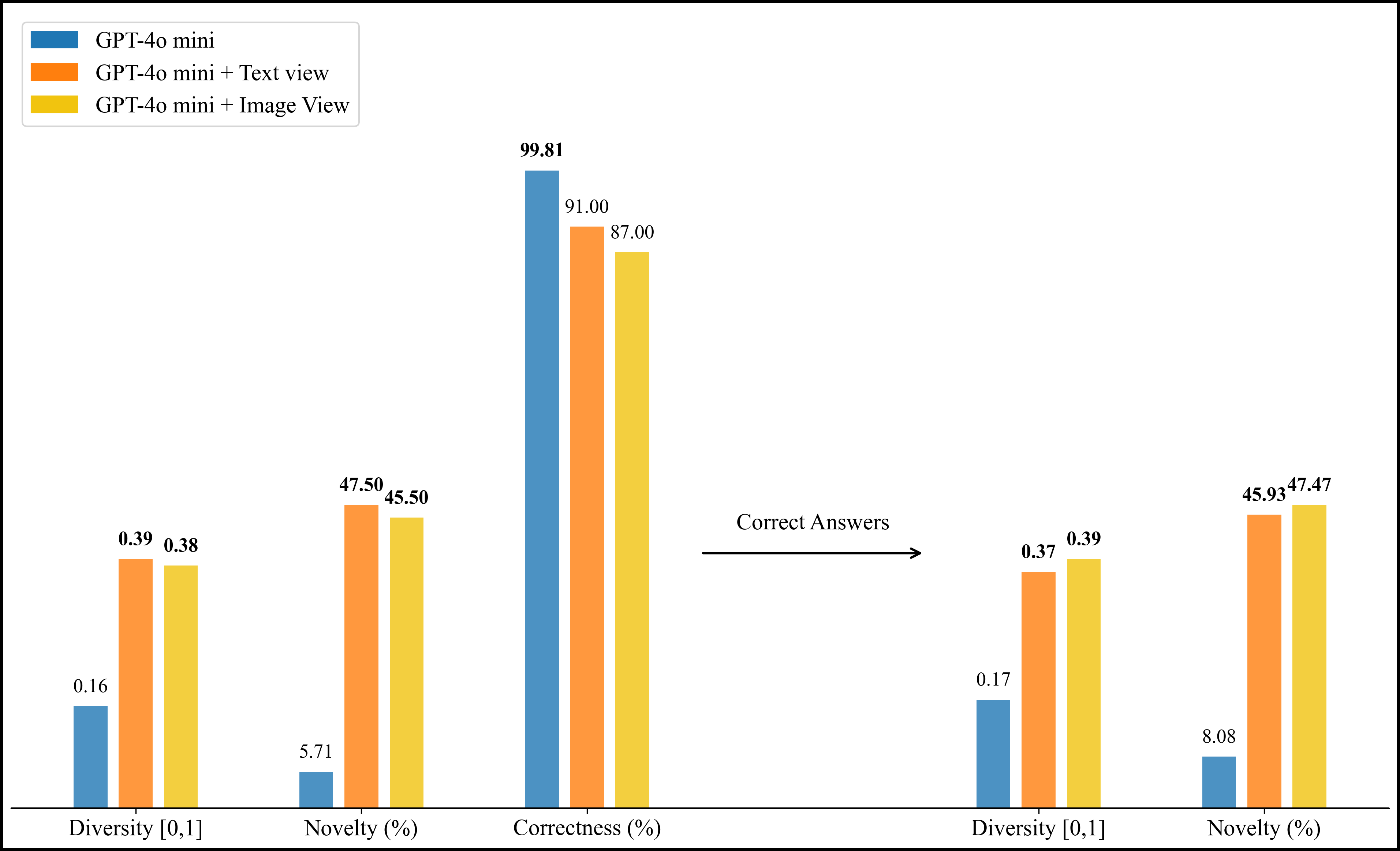} \hspace{3em}
  \includegraphics[width=0.45\linewidth]{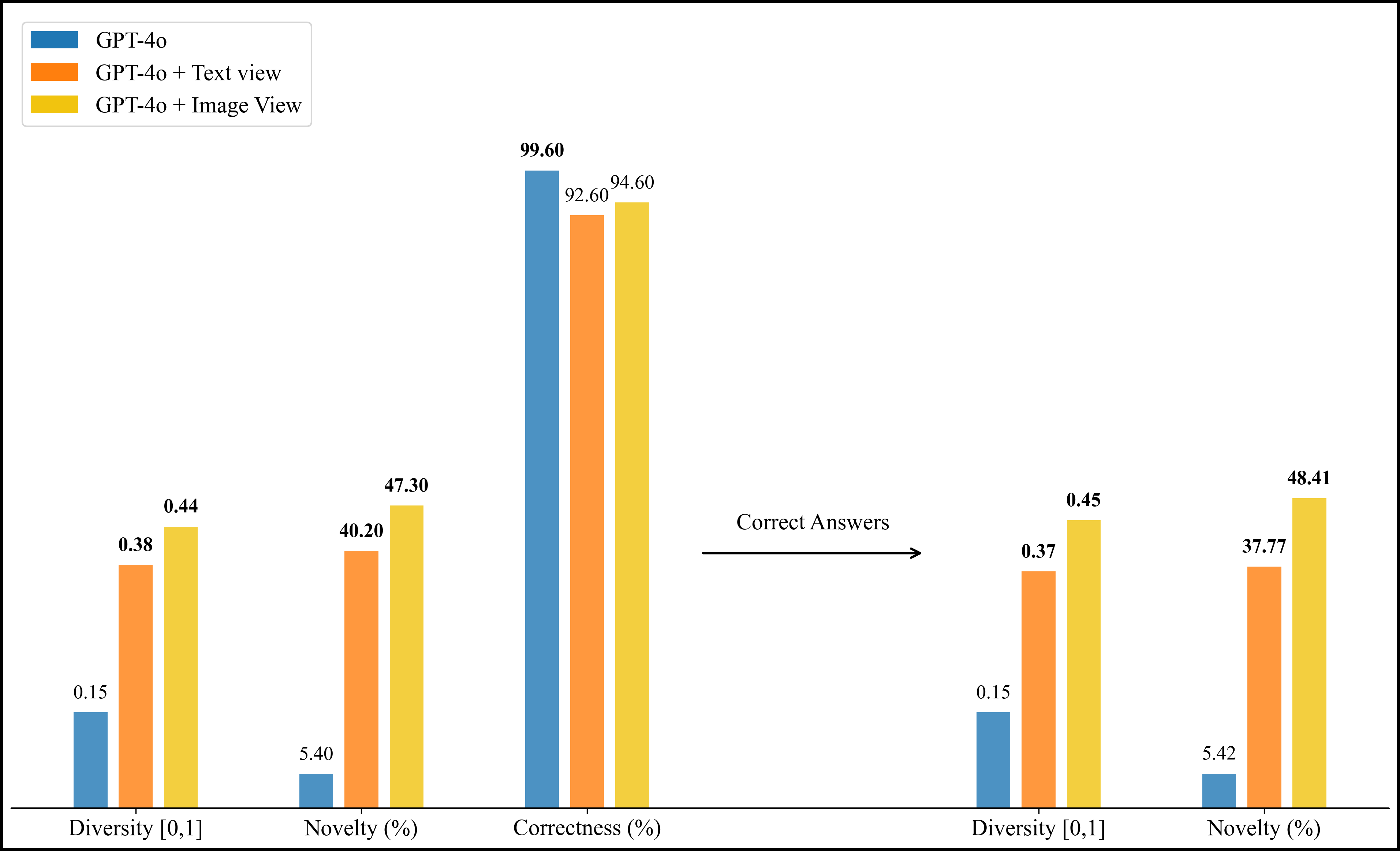}
  \caption{Diversity, Novelty, and Correctness (All answers) vs (Correct answers) For $num\_samples = 100$. Diversity measure is $Self\_BLUE$, Novelty measure is based on $SBERT$, and Correctness measure is based on $GPT-4o$.}
  \label{fig:final_comparison}
\end{figure*}

The evaluation of answer correctness across different models is presented in Table \ref{tab:correctness_combined_single}, considering two aspects introduced in Section \ref{subsubsec: correctness_measure}. We assess the correctness of generated answers based on their relevance to the input prompt, using GPT-4o as the correctness evaluator and evaluates the correctness of language models from an English language structure perspective with DeepSeekV3. The results indicate that all models achieve high correctness scores in this aspect. Figure \ref{fig:final_comparison} summarizes the performance of our proposed method (multi-view embedding) alongside standard model variants. We report three key metrics: diversity, novelty, and correctness. Diversity quantifies the variety of generated responses on scale of [0,1], novelty measures the proportion of novel content (expressed in percentages), and correctness reflects the relevance of the generated text to the given prompt.
Overall, the data show that applying our method increases both diversity and novelty across various models. For instance, with Qwen, the baseline achieves a diversity of 0.41, novelty of 32.7\%, and correctness of 93.77\%. When we apply the text-view, diversity rises to 0.56 and novelty to 68.7\%, albeit with a correctness drop to 76.6\%. Similarly, the image-view offers 0.50 diversity and 57.2\% novelty at a correctness of 82.5\%. These patterns exemplify a trade-off: while the generated content becomes more varied and inventive, alignment with the prompt can decrease.
To further evaluate whether these gains hold when only correct (i.e., relevant) answers are considered, we also compute diversity and novelty exclusively for the subset of correct responses. Even under this stricter measurement, the Qwen variants (text view and image view) maintain higher diversity and novelty than the baseline. This finding underscores that our method fosters creativity and variety without entirely sacrificing correctness..

\begin{table}[t] 
\centering
\resizebox{\linewidth}{!}{%
\begin{tabular}{lcc}
\toprule
 & \multicolumn{2}{c}{\textbf{Correctness Measure}} \\
\cmidrule(lr){2-3}
\textbf{Model} & \textbf{Correctness (\%)} & \textbf{Language Structure ([1,10])} \\
\midrule
GPT-2 & 7.20 & 3.29 \\
GPT-2 + Text View & 2.40 & 2.40 \\
\hline
Qwen & 93.77 & 7.57 \\
Qwen + Text View & 76.60 & 7.52 \\
Qwen + Image View & 82.50 & 6.64 \\
\hline
LlaMa-3.2 & 85.60 & 8.42 \\
LlaMa-3.2 + Text View & 83.10 & 7.38 \\
LlaMa-3.2 + Image View & 90.30 & 7.79 \\
\hline
DeepSeek-R1 & 91.80 & 7.96 \\
DeepSeek-R1 + Text View & 81.00 & 7.15 \\
DeepSeek-R1 + Image View & 53.90 & 6.10 \\
\hline
GPT-4o mini & 99.81 & 8.70 \\
GPT-4o mini + Text View & 91.00 & 8.12 \\
GPT-4o mini + Image View & 87.00 & 7.71 \\
\hline
GPT-4o & 99.60 & 8.07 \\
GPT-4o + Text View & 92.60 & 8.05 \\
GPT-4o + Image View & 94.60 & 8.06 \\
\bottomrule
\end{tabular}%
}
\caption{Correctness (\%) based on GPT-4o and language structure based on DeepSeekV3 ([1,10]). $num\_sample =100$ per each prompt (10 prompts).}
\label{tab:correctness_combined_single}
\end{table}

\section{Conclusion}
In this work, we present multi-view embedding, a model-agnostic approach that enriches the input prompt with diverse textual and visual sources to enhance the diversity and novelty of generated responses for both open-source and closed-source LLMs. Additionally, we introduce a framework to assess the diversity, novelty, and correctness of the generated outputs from LLMs. We validate both our method and framework using real-world datasets and $909k$ generated answers from various LLMs.

\section{Future Work}
As demonstrated in this paper, incorporating textual or visual views enhances the output diversity and creativity of existing language models. In our future work, we aim to explore the impact of additional view types, such as videos, songs, maps, or even mathematical equations, as multi-view embeddings on model performance. Furthermore, we plan to extend our proposed framework to assess not only diversity, novelty, and correctness, but also effectiveness and usefulness, in order to provide a more comprehensive evaluation of the behavior of large language models.

\clearpage
\clearpage
\section{Limitations} 
Despite the promising results, our method heavily depends on the quality and relevance of the views used during generation. If text or image views are not closely related to the input prompt, the model may produce outputs with lower correctness and reduced novelty. Additionally, generating these views—particularly in large numbers—can itself pose challenges. In our work, we use GPT-4o as a multi-view text generator; however, if many distinct views are requested, their diversity and novelty may diminish. This means that even if the underlying mechanism can foster creativity, oversaturating the process with a high volume of views can lead to repetitive or less innovative content. Finally, while our approach can enhance creativity and diversity for a moderate number of outputs, producing a large number of responses inevitably reduces both novelty and variation. These factors suggest that our method is most effective for situations where a bounded number of creative yet coherent outputs is desired, rather than an unbounded volume of responses.

 \bibliography{acl_latex}

\appendix
\section{Appendix: Detailed Experimental Results}

\label{appendix:a}
\begin{figure}[ht!]
    \centering
    \noindent
    \begin{minipage}{0.35\textwidth} 
        \includegraphics[width=\textwidth]{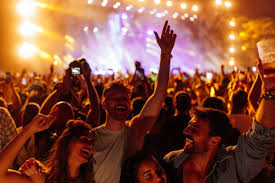} 
    \end{minipage}%
    \hspace{1em} 
    \begin{minipage}{0.4\textwidth}
        \vspace{1em} 
        \justifying 
        \noindent\textbf{Prompt:} \textit{What is the meaning of true happiness in life?} \\
        \noindent\textbf{Image View:} \textit{The scene shows a vibrant outdoor concert where a large crowd is gathered, with many people expressing their enjoyment by raising their hands in the air. Bright lights illuminate the area, particularly focusing on the stage in the background. The audience is lively and energetic, with some individuals recording the experience on their phones. The atmosphere is filled with excitement and celebration, characteristic of a live music event.}
    \end{minipage}

    \vspace{1em} 
    \justifying
    \noindent\textbf{Answer:} \textit{True happiness in life often encompasses experiences that bring joy, connection, and personal growth through various activities such as participating in music concerts like the one described. Engaging in shared experiences with others creates bonds and enriches our social lives. Music, being an art form rooted deeply in human culture, has long been associated with positive emotions; it can evoke strong feelings of joy, community, and self-expression. In such a vibrant outdoor concert setting, the collective emotion displayed by the attendees—raising their hands in the air, indicating enthusiasm, and enthusiastically recording the moment with smartphones—are vivid expressions of true happiness and engagement.}

    \caption{An example of Image view related to the input prompt and answered by Qwen2.5-1.5B model.}
    \label{fig:happiness}
\end{figure}

\begin{table}[ht]
    \centering
    \resizebox{\linewidth}{!}{ 
    \begin{tabular}{ l c c c c c }
        \hline
        \textbf{Models} & \textbf{MTLD} & \textbf{TF-IDF} & \textbf{Lexical Entropy} & \textbf{Semantic Embedding} & \textbf{Self-BLEU} \\
        \hline
        GPT-2                      & \textbf{41.44}  & 0.85  & 7.98  & 0.54  & 0.76  \\
    
        GPT-2 + Text View          & 35.91  & \textbf{0.88}  & \textbf{8.11}  & \textbf{0.64}  & \textbf{0.80}  \\
        \hline
        Qwen-2.5                       & 67.00  & 0.78  & 7.81  & 0.22  & 0.41  \\
        
        Qwen-2.5 + Text View           & 67.77  & \textbf{0.85}  & \textbf{8.32}  & \textbf{0.39}  & \textbf{0.56}  \\
        
        Qwen-2.5 + Image View          & \textbf{70.24}  & 0.82  & 8.15  & 0.29  & 0.50  \\
        \hline
        LLama-3.2                  & 52.20 & 0.70 & 7.32 & 0.23 & 0.31\\
        LLama-3.2 + Text View      & 54.22 & \textbf{0.82} & \textbf{8.04} & \textbf{0.44} & \textbf{0.54}\\
        LLama-3.2 + Image View     & \textbf{54.39} & 0.74 & 7.59 & 0.31 & 0.41\\
        \hline
        DeepSeek-R1             & 52.34  & 0.76  & 7.54  & 0.24  & 0.39  \\
        
        DeepSeek-R1 + Text View & \textbf{55.07}  & \textbf{0.84}  & \textbf{8.07}  & \textbf{0.40}  & \textbf{0.52}  \\
        
        DeepSeek-R1 + Image View & 54.37  & 0.82  & 7.95  & 0.39  & \textbf{0.52}  \\
        \hline
        GPT-4o Mini                & 57.98  & 0.61  & 7.03  & 0.11  & 0.16  \\
        
        GPT-4o Mini + Text View    & 59.68  & \textbf{0.77}  & \textbf{7.87}  & \textbf{0.34}  & \textbf{0.39}  \\
        
        GPT-4o Mini + Image View   & \textbf{60.19}  & \textbf{0.77}  & 7.76  & 0.32  & 0.38  \\
        \hline
        GPT-4o                     & 57.17  & 0.64  & 6.85  & 0.10  & 0.15  \\
        
        GPT-4o + Text View         & \textbf{58.25}   & 0.76  & \textbf{7.75}  & \textbf{0.31}  & 0.38  \\
        
        GPT-4o + Image View        & 54.66   & \textbf{0.77}  & 7.68  & \textbf{0.31}  & \textbf{0.44}  \\
        \hline
        
    \end{tabular}
    } 
    \caption{Diversity Metrics for different models. Higher values indicate better performance. $num\_samples = 100$ , $max\_length = 125$}
    \label{tab:num_samples100}
\end{table}

\begin{table}[ht]
    \centering
    \resizebox{\linewidth}{!}{ 
    \begin{tabular}{ l c c c c c }
        \hline
        \textbf{Models} & \textbf{MTLD} & \textbf{TF-IDF} & \textbf{Lexical Entropy} & \textbf{Semantic Embedding} & \textbf{Self-BLEU} \\
        \hline
        GPT-2                      & \textbf{41.35}  & 0.86  & 8.11  & 0.55  & 0.68  \\
    
        GPT-2 + Text View          & 35.69  & \textbf{0.89}  & \textbf{8.28}  & \textbf{0.64}  & \textbf{0.72}  \\
        \hline
        Qwen-2.5                       & 67.45  & 0.79  & 7.88  & 0.22  & 0.31  \\
        
        Qwen-2.5 + Text View           & 67.68  & \textbf{0.85}  & \textbf{8.43}  & \textbf{0.39}  & 0.47  \\
        
        Qwen-2.5 + Image View          & \textbf{70.27}  & 0.83  & 8.26  & 0.30  & \textbf{0.48}  \\
        \hline
        LLama-3.2                  & 52.52 & 0.71 & 7.38 & 0.22 & 0.23\\
        LLama-3.2 + Text View      & 54.31 & \textbf{0.83} & \textbf{8.16} & \textbf{0.43} & \textbf{0.44}\\
        LLama-3.2 + Image View     & \textbf{54.59} & 0.75 & 7.66 & 0.30 & 0.31\\
        \hline
        DeepSeek-R1             & 52.41  & 0.78  & 7.60  & 0.24  & 0.30  \\
        
        DeepSeek-R1 + Text View & \textbf{54.44}  & \textbf{0.85}  & \textbf{8.12}  & \textbf{0.40}  & \textbf{0.45}  \\
        
        DeepSeek-R1 + Image View & 54.42  & 0.83  & 8.05  & 0.39  & 0.41  \\
        \hline
        GPT-4o Mini                & 57.94  & 0.62  & 7.05  & 0.11  & 0.16  \\
        
        GPT-4o Mini + Text View    & 59.79  & \textbf{0.79}  & \textbf{7.91}  & \textbf{0.33}  & \textbf{0.38}  \\
        
        GPT-4o Mini + Image View   & \textbf{60.38}  & 0.78  & 7.82  & 0.31  & 0.28  \\
        \hline
        GPT-4o                     & 57.40  & 0.66  & 6.88  & 0.09  & 0.10  \\
        
        GPT-4o + Text View         & \textbf{58.21}   & 0.78  & \textbf{7.85}  & \textbf{0.31}  & 0.29  \\
        
        GPT-4o + Image View        & 55.23   & \textbf{0.79}  & 7.77  & \textbf{0.31}  & \textbf{0.34}  \\
        \hline
        
    \end{tabular}
    } 
    \caption{Diversity Metrics for different models. Higher values indicate better performance. $num\_samples = 250$, $max\_length = 125$.}
    \label{tab:num_sample250}
\end{table}

\begin{table}[ht]
    \centering
    \resizebox{\linewidth}{!}{ 
    \begin{tabular}{ l c c c c c }
        \hline
        \textbf{Models} & \textbf{MTLD} & \textbf{TF-IDF} & \textbf{Lexical Entropy} & \textbf{Semantic Embedding} & \textbf{Self-BLEU} \\
        \hline
        GPT-2                      & \textbf{41.30}  & 0.87  & 8.17  & 0.55  & 0.60  \\
    
        GPT-2 + Text View          & 35.77  & \textbf{0.90}  & \textbf{8.36}  & \textbf{0.64}  & \textbf{0.65}  \\
        \hline
        Qwen-2.5                       & 67.13  & 0.80  & 7.91  & 0.22  & 0.25  \\
        
        Qwen-2.5 + Text View           & 67.62  & \textbf{0.86}  & \textbf{8.48}  & \textbf{0.39}  & \textbf{0.40}  \\
        
        Qwen-2.5 + Image View          & \textbf{70.36}  & 0.84  & 8.32  & 0.30  & 0.36  \\
        \hline
        LLama-3.2                  & 52.60 & 0.72 & 7.43 & 0.23 & 0.18\\
        LLama-3.2 + Text View      & 54.22 & \textbf{0.84} & \textbf{8.23} & \textbf{0.43} & \textbf{0.37}\\
        LLama-3.2 + Image View     & \textbf{54.45} & 0.76 & 7.71 & 0.31 & 0.26\\
        \hline 
        DeepSeek-R1             & 52.56  & 0.79  & 7.66  & 0.24  & 0.24  \\
        
        DeepSeek-R1 + Text View & \textbf{54.52}  & \textbf{0.86}  & \textbf{8.22}  & \textbf{0.39}  & \textbf{0.37}  \\
        
        DeepSeek-R1 + Image View & 54.27  & 0.84  & 8.08  & \textbf{0.39}  & 0.34  \\
        \hline
        GPT-4o Mini                & 57.95  & 0.63  & 7.07  & 0.11  & 0.07  \\
        
        GPT-4o Mini + Text View    & 59.53  & \textbf{0.79}  & \textbf{7.93}  & \textbf{0.33}  & \textbf{0.22}  \\
        
        GPT-4o Mini + Image View   & \textbf{60.36}  & \textbf{0.79}  & 7.85  & 0.31  & 0.21  \\
        \hline
        GPT-4o                     & 56.99  & 0.67  & 6.89  & 0.09  & 0.07  \\
        
        GPT-4o + Text View         &  \textbf{58.22}  &  \textbf{0.79} & \textbf{7.86}  & 0.30  & 0.23  \\
        
        GPT-4o + Image View        & 55.29   & 0.79  & \textbf{7.79}  & \textbf{0.31}  & \textbf{0.27}  \\
        \hline
        
    \end{tabular}
    } 
    \caption{Diversity Metrics for different models. Higher values indicate better performance. $num\_samples = 500$, $max\_length = 125$.}
    \label{tab:num_samples500}
\end{table}

\begin{table}[ht]
    \centering
    \resizebox{\linewidth}{!}{ 
    \begin{tabular}{ l c c c c c }
        \hline
        \textbf{Models} & \textbf{MTLD} & \textbf{TF-IDF} & \textbf{Lexical Entropy} & \textbf{Semantic Embedding} & \textbf{Self-BLEU} \\
        \hline
        GPT-2                      & \textbf{41.33}  & 0.87  & 8.21  & 0.54  & 0.53  \\
    
        GPT-2 + Text View          & 35.66  & \textbf{0.90}  & \textbf{8.43}  & \textbf{0.64}  & \textbf{0.59}  \\
        \hline
        Qwen-2.5                       & 67.10  & 0.81  & 7.94  & 0.22  & 0.20  \\
        
        Qwen-2.5 + Text View           & 67.44  & \textbf{0.87}  & \textbf{8.50}  & \textbf{0.38}  & \textbf{0.34}  \\
        
        Qwen-2.5 + Image View          & \textbf{70.29}  & 0.85  & 8.34  & 0.30  & 0.30  \\
        \hline
        LLama-3.2                  & 52.63 & 0.73 & 7.44 & 0.23 & 0.14\\
        LLama-3.2 + Text View      & 54.22 & \textbf{0.84} & \textbf{8.26} & \textbf{0.43} & 0.31\\
        LLama-3.2 + Image View     & \textbf{54.43} & 0.77 & 7.74 & 0.31 & \textbf{0.21}\\
        \hline
        DeepSeek-R1             & 52.54  & 0.79  & 7.67  & 0.24  & 0.19  \\
        
        DeepSeek-R1 + Text View & \textbf{54.55}  & \textbf{0.86}  & \textbf{8.27}  & \textbf{0.39}  & \textbf{0.31}  \\
        
        DeepSeek-R1 + Image View & 54.38  & 0.84  & 8.11  & \textbf{0.39}  & 0.28  \\
        \hline
        GPT-4o Mini                & 57.95  & 0.64  & 7.08  & 0.11  & 0.05  \\
        
        GPT-4o Mini + Text View    & \textbf{59.57}  & \textbf{0.88}  & \textbf{7.95}  & \textbf{0.33}  & 0.17  \\
        
        GPT-4o Mini + Image View   & 55.76  & 0.81  & 7.79  & 0.29  & \textbf{0.19}  \\
        \hline
        GPT-4o                     & 57.83  & 0.66  & 7.14  & 0.13  & 0.06  \\
        
        GPT-4o + Text View         & \textbf{58.29}   & 0.79  & \textbf{7.90}  & \textbf{0.31}  & 0.18  \\
        
        GPT-4o + Image View        & 55.12   & \textbf{0.80}  & 7.80  & \textbf{0.31}  & \textbf{0.21}  \\
        \hline
        
    \end{tabular}
    } 
    \caption{Diversity Metrics for different models. Higher values indicate better performance. $num\_samples = 1000$, $max\_length = 125$.}
    \label{tab:num_samples1000}
\end{table}

\begin{table}[ht]
    \centering
    \resizebox{\linewidth}{!}{ 
    \begin{tabular}{ l c c c c c }
        \hline
        \textbf{Models} & \textbf{MTLD} & \textbf{TF-IDF} & \textbf{Lexical Entropy} & \textbf{Semantic Embedding} & \textbf{Self-BLEU} \\
        \hline
        GPT-2                      & \textbf{41.39}  & 0.87  & 8.24  & 0.55  & 0.49  \\
    
        GPT-2 + Text View          & 33.30  & \textbf{0.90}  & \textbf{8.47}  & \textbf{0.66}  & \textbf{0.58}  \\
        \hline
        Qwen-2.5                       & 67.19  & 0.81  & 7.95  & 0.22  & 0.18  \\
        
        Qwen-2.5 + Text View           & 67.64   & \textbf{0.87}  & \textbf{8.52}  & \textbf{0.38}  & \textbf{0.30}  \\

        Qwen-2.5 + Image View          & \textbf{70.30}  & 0.85  & 8.35  & 0.30  & 0.27  \\
        \hline
        LLama-3.2                  & 52.53 & 0.73 & 7.44 & 0.23 & 0.12\\
        LLama-3.2 + Text View      & 54.28 & \textbf{0.85} & \textbf{8.28} & \textbf{0.43} & \textbf{0.27}\\
        LLama-3.2 + Image View     & \textbf{54.40} & 0.77 & 7.75 & 0.31 & 0.18\\
        \hline
        DeepSeek-R1             & \textbf{52.59}  & 0.79  & 7.68  & 0.24  & 0.17  \\
        
        DeepSeek-R1 + Text View & \textbf{54.53}  & \textbf{0.86}  & \textbf{8.28}  & \textbf{0.39}  & \textbf{0.28}  \\
        
        DeepSeek-R1 + Image View & 54.32  & 0.84  & 8.12  & \textbf{0.39}  & 0.24  \\
        \hline
        GPT-4o Mini                & 57.91  & 0.64  & 7.08  & 0.11  & 0.04  \\
        
        GPT-4o Mini + Text View    & \textbf{59.50}  & \textbf{0.80}  & \textbf{7.96}  & \textbf{0.33}  & 0.15  \\
        
        GPT-4o Mini + Image View   & 54.86  & 0.81  & 7.77  & 0.29  & \textbf{0.16}  \\
        \hline
        GPT-4o                     & 57.90  & 0.66  & 7.14  & 0.13  & 0.05  \\
        
        GPT-4o + Text View         & \textbf{58.26}   & 0.79  & \textbf{7.90}  & \textbf{0.31}  & 0.15  \\
        
        GPT-4o + Image View        & 55.14   & \textbf{0.80}  & 7.82  & \textbf{0.31}  & \textbf{0.18}  \\
        \hline
        
    \end{tabular}
    } 
    \caption{Diversity Metrics for different models. Higher values indicate better performance. $num\_samples = 1500$, $max\_length = 125$.}
    \label{tab:num_samples1500}
\end{table}
\begin{table}[ht]
    \centering
    \resizebox{\linewidth}{!}{ 
    \begin{tabular}{ l c c c c c }
        \hline
        \textbf{Models} & \textbf{MTLD} & \textbf{TF-IDF} & \textbf{Lexical Entropy} & \textbf{Semantic Embedding} & \textbf{Self-BLEU} \\
        \hline
        GPT-2                      & \textbf{41.36}  & 0.87  & 8.25  & 0.55  & 0.46  \\
    
        GPT-2 + Text View          & 33.39  & \textbf{0.90}  & \textbf{8.48}  & \textbf{0.67}  & \textbf{0.55}  \\
        \hline
        Qwen-2.5                       & 67.25 & 0.81  & 7.95  & 0.22  & 0.16  \\
        
        Qwen-2.5 + Text View           & 67.68   & \textbf{0.87}  & \textbf{8.53}  & \textbf{0.38}  & \textbf{0.28}  \\

        Qwen-2.5 + Image View          & \textbf{70.24}  & 0.85  & 8.36  & 0.30  & 0.25  \\
        \hline
        LLama-3.2                  & 52.21 & 0.73 & 7.45 & 0.23 & 0.11\\
        LLama-3.2 + Text View      & 54.14 & \textbf{0.85} & \textbf{8.29} & \textbf{0.43} & \textbf{0.25}\\
        LLama-3.2 + Image View     & \textbf{54.49} & 0.77 & 7.75 & 0.30 & 0.16\\
        \hline
        DeepSeek-R1             & 52.62  & 0.79  & 7.69  & 0.24  & 0.15  \\
        
        DeepSeek-R1 + Text View & \textbf{54.53}  & \textbf{0.86}  & \textbf{8.29}  & \textbf{0.39} & \textbf{0.26}  \\
        
        DeepSeek-R1 + Image View & 54.38  & 0.84  & 8.12  & \textbf{0.39}  & 0.22  \\
        \hline
        GPT-4o Mini                & 57.98  & 0.64  & 7.08  & 0.11  & 0.03  \\
        
        GPT-4o Mini + Text View    & \textbf{59.50}  & 0.80  & \textbf{7.97}  & \textbf{0.33}  & 0.13  \\
        
        GPT-4o Mini + Image View   & 54.90  & \textbf{0.81}  & 7.78  & 0.29  & \textbf{0.14}  \\
        \hline
        GPT-4o                     & 55.78  & 0.67  & 7.16  & 0.12  & 0.05  \\
        
        GPT-4o + Text View         & 58.26   & \textbf{0.80}  & 7.91  & 0.31  & 0.14  \\
        
        GPT-4o + Image View        &\textbf{59.05}   & \textbf{0.80}  & \textbf{7.94}  & \textbf{0.35}  & \textbf{0.15}  \\
        \hline
        
    \end{tabular}
    } 
    \caption{Diversity Metrics for different models. Higher values indicate better performance. $num\_samples = 2000$, $max\_length = 125$.}
    \label{tab:num_samples2000}
\end{table}

\begin{table}[ht]
    \centering
    \resizebox{\linewidth}{!}{ 
    \begin{tabular}{ c c c c c c}
        \hline
        \textbf{Seed} & \textbf{Model} & \textbf{Accuracy} & \textbf{Precision} & \textbf{Recall} & \textbf{F1-score} \\
        \hline
        \multirow{2}{*}{Seed0}  & DeepSeekV3  & 0.8450 & \textbf{0.9367} & 0.7400 & 0.8268 \\
                                & GPT-4o       & \textbf{0.9100} & 0.9271 & \textbf{0.8900} & \textbf{0.9082} \\
        \hline
        \multirow{2}{*}{Seed1}  & DeepSeekV3  & 0.8100 & \textbf{0.9198} & 0.6800 & 0.7816 \\
                                & GPT-4o       & \textbf{0.8700} & 0.9111 & \textbf{0.8200} & \textbf{0.8632} \\
        \hline
        \multirow{2}{*}{Seed2}  & DeepSeekV3  & 0.8450 & 0.9726 & 0.7100 & 0.8187 \\
                                & GPT-4o       & \textbf{0.9150} & \textbf{0.9663} & \textbf{0.8600} & \textbf{0.9091} \\
        \hline
        \multirow{2}{*}{Seed3}  & DeepSeekV3  & 0.7900 & 0.9265 & 0.6300 & 0.7500 \\
                                & GPT-4o       & \textbf{0.8650} & \textbf{0.9195} & \textbf{0.8000} & \textbf{0.8556} \\
        \hline
        \multirow{2}{*}{Seed4}  & DeepSeekV3  & 0.8750 & 0.9000 & 0.7600 & 0.8588 \\
                                & GPT-4o       & \textbf{0.9300} & \textbf{0.9778} & \textbf{0.8800} & \textbf{0.9268} \\
        \hline
        \multirow{2}{*}{Seed5}  & DeepSeekV3  & 0.8300 & 0.9342 & 0.7200 & 0.8068 \\
                                & GPT-4o       & \textbf{0.9200} & \textbf{0.9773} & \textbf{0.8600} & \textbf{0.9149} \\
        \hline
        \multirow{2}{*}{Seed6}  & DeepSeekV3  & 0.8450 & \textbf{0.9859} & 0.7000 & 0.8187 \\
                                & GPT-4o       & \textbf{0.8650} & 0.9011 & \textbf{0.8000} & \textbf{0.8556} \\
        \hline
        \multirow{2}{*}{Seed7}  & DeepSeekV3  & 0.8550 & \textbf{0.9863} & 0.7200 & 0.8324 \\
                                & GPT-4o       & \textbf{0.9150} & 0.9462 & \textbf{0.8800} & \textbf{0.9123} \\
        \hline
        \multirow{2}{*}{Seed8}  & DeepSeekV3  & 0.8400 & \textbf{0.9857} & 0.6900 & 0.8118 \\
                                & GPT-4o       & \textbf{0.8850} & 0.9753 & \textbf{0.7900} & \textbf{0.8729} \\
        \hline
        \multirow{2}{*}{Seed9}  & DeepSeekV3  & 0.8450 & \textbf{0.9600} & 0.7200 & 0.8229 \\
                                & GPT-4o       & \textbf{0.8750} & 0.9032 & \textbf{0.8400} & \textbf{0.8750} \\
        \hline
    \end{tabular}
    } 
    \caption{Classification metricss for GPT-4o and DeepSeekV3 across different seeds to evaluate the ability of these models to detecet the correct answer.}
    \label{tab:classification_metrics_correctness}
\end{table}

\begin{table}
\centering

\resizebox{\linewidth}{!}{ 
\begin{tabular}{lccc}
\toprule
 & \multicolumn{1}{c}{num\_samples = 100} & \multicolumn{1}{c}{num\_samples = 250} & \multicolumn{1}{c}{num\_samples = 500} \\
\cmidrule(lr){2-2} \cmidrule(lr){3-3} \cmidrule(lr){4-4}
Model & SBERT & SBERT & SBERT \\
\midrule
GPT2 & 98.10 & 96.76 & 95.46 \\
GPT2 + Text View & 99.70 & 98.88 & 98.72 \\
\hline
GPT-4o-mini & 5.71 & 3.44 & 2.50 \\
GPT-4o-mini + Text View & 47.50 & 32.24 & 24.30 \\
GPT-4o-mini + Image View & 47.30 & 30.44 & 18.70 \\
\hline
Qwen2.5-1.5B & 32.70 & 25.44 & 22.58 \\
Qwen2.5-1.5B + Text View & 68.70 & 59.60 & 52.64 \\
Qwen2.5-1.5B + Image View & 57.20 & 48.80 & 45.14 \\
\bottomrule
\end{tabular}}
\caption{Results on percentage of novelty score across different models according to the novelty detector SBERT}
\label{tab:novelty_sbert_2}
\end{table}

\clearpage
\noindent\textbf{Example for correct and incorrect answer:}\label{sec:correctness_measure_example}\\
The generated answers to the input prompt could be novel, or even diverse but incorrect. Pay attention to the following example:.\\
{\ttfamily
\justifying
Prompt: "what is the meaning of true happiness in life?"\\
Answer 1: (First answer$\rightarrow$ novel, correct)\\
"True happiness in life is achieving a balance between personal fulfillment and meaningful connections with others, along with good health and peace of mind."\\
Answer 2: (novel, and correct)\\
"Happiness means being content with what you have, staying optimistic in the face of challenges, and cherishing the moments shared with family and friends."\\
Answer 3: (non-novel, and correct)\\
“Happiness is living a healthy life, having supportive relationships, achieving your goals, and feeling grateful for what you have.”\\
Answer 4: (novel, but incorrect)\\
"To make the best pizza dough, you need high-quality flour, water, yeast, and a bit of olive oil. Let it rise for at least 24 hours for optimal texture."\\

\par}

\section{Appendix: Prompts and more information}
\label{appendix:b}

\begin{figure}[ht!]
    \centering
    \includegraphics[width=0.55\textwidth]{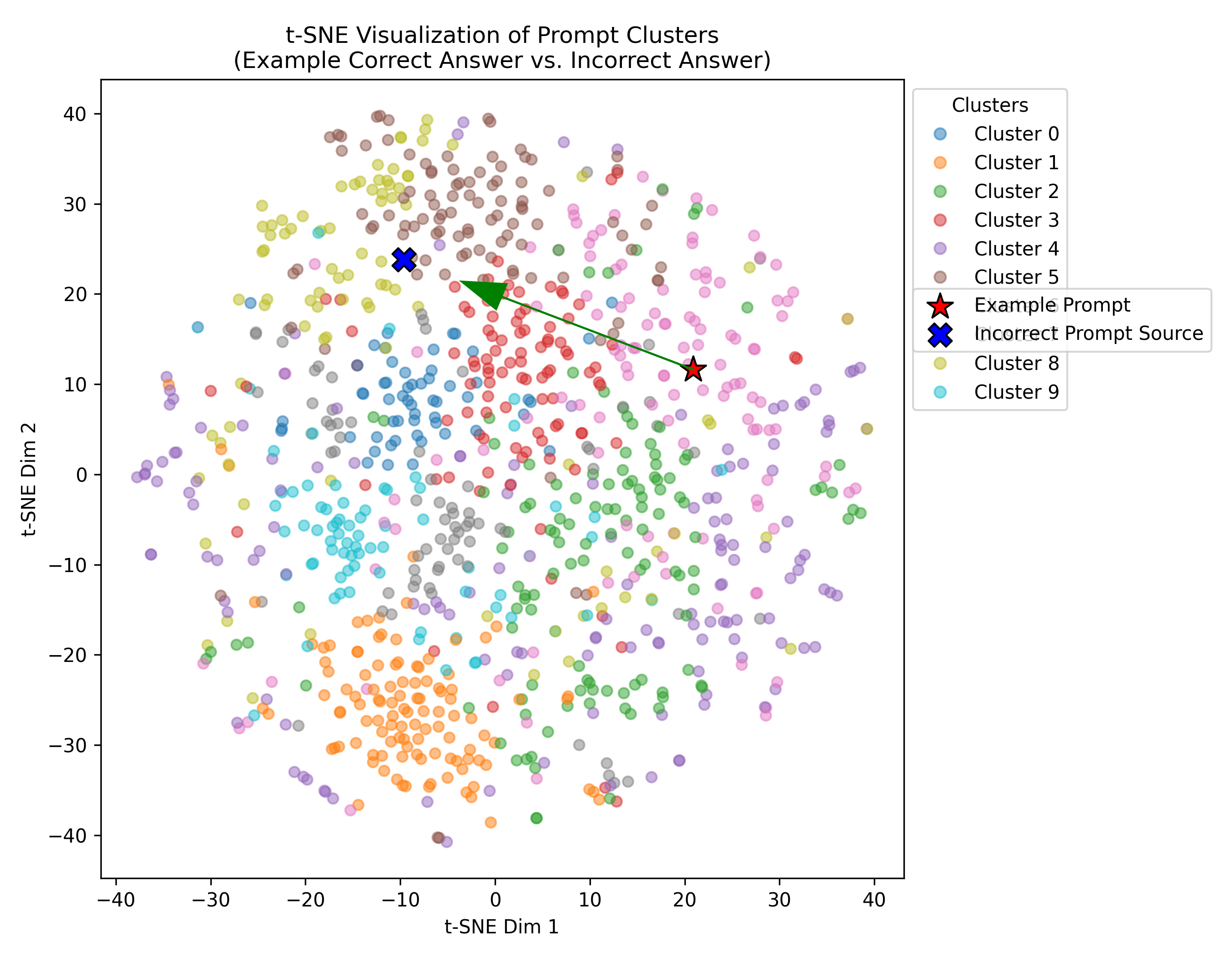} 
    \caption{An example of selecting correct and incorrect prompt for a prompt and summarized answers.}
    \label{fig:prompt_answer_correctness}
\end{figure}

\noindent\textbf{Prompts used in this work}\\
We Used 10 prompts from different subjects to improve generalization on our results.\\
{\ttfamily
\justifying
1.Philosophical Question:\\
"What is the meaning of true happiness in life?",\\

2.Hypothetical Scenario:\\
"If humans could live on Mars, what challenges would they face and how could they overcome them?",\\

3.Creative Thinking Prompt:\\
"Can you describe an imaginary city where technology and nature exist in perfect harmony?",\\

4.Practical Advice Question:\\
"What are the most effective ways to learn a new language quickly?",\\

5.Exploration of Abstract Concepts:\\
"How would you explain the concept of time to someone who has never experienced it?",\\
    
6.Scientific Exploration:\\
"What are the possible effects of artificial intelligence on scientific research in the next decade?",\\

7.Ethical Dilemma:\\
"Is it ever justifiable to prioritize technological advancement over environmental protection?",\\

8.Problem-Solving Question:\\
"How can cities effectively reduce traffic congestion without compromising accessibility?",\\

9.Imaginative Scenario:\\
"If animals could communicate with humans, how would that change our world?",\\

10.Personal Reflection Prompt:\\
"What qualities make someone a great leader, and how can those qualities be developed?"\\
    
\par}
\noindent\textbf{Evaluate the correctness of prompt answer experiment details} \\
This is an example of a random prompt from seed 0 with it's correct answer and selected incorrect answer from another cluster. in Figure \ref{fig:prompt_answer_correctness} the visualization of distance between two answers demonstrated.\\

{\ttfamily
\justifying
Prompt:\\
"In a isolated research facility , a researcher walks past a janitor , only to stop later and realize that we do n't have a janitor ."\\
"correct\_answer":\\
"Something was off . But something was always off in these hallways . The corridors slightly too long , slightly too narrow for two people to walk comfortably abreast . As if someone had googled exactly what it was that made people unconsciously uncomforTable and crammed every last wretched detail into the place . Shifting in the starched collar , he sniffed and pulled the pen from the clipboard . <newline> Trying to trace the events that lead to him being here , it was nigh on impossible . A odd comment by a professor there , reading a specific page of a newspaper , there was no defining reason that spurred him here . <newline> That said , no one could deny that it was a bold career move , and a bold move for him in general . He was in Antarctica for Christ 's sake . <newline> It brought him no small joy that on slow days he would pace the corridors of the facility , for lack of something better to do . He had worked out the optimal route based on path efficiency , and had divided his routine between Surveying tasks so that the laps loosely resembled a set - three reps of the uniformly grey corridors , spaced by intervals of 45 minutes . <newline> So on the second rep of the routine , passing through a adjunct corridor connecting both polar fieldwork hub A and a room masquerading as a laboratory , that he saw a Figure that the very end of the hallway . <newline> It 's not my job to speculate on a person 's intelligence obviously . But what I can say is it was a pretty damn long corridor and our protagonist had enough time to realise that something was off . Instead , the approach happened , he adopted his socially trained response to the presence of an unknown but nonthreatening human , an easy smile , and nodded as he went past the scruffy Figure . <newline> He took in some details . The janitors uniform , one that looked from the lack of utility more parodic that anything , as if dressed Halloween or something . A slight gait , no hat , two buttons done up interrupted by the third undone and then the fourth one buttoned up by the neck . <newline> He rounded the corner . <newline> There was a pause . <newline> He reprocessed the moment . <newline> Turning , he popped his head back around the corner , and looked back down the corridor , visibly straining . <newline> 'Mom ? ' <newline>"\\
"incorrect\_answer":\\
"*God is everywhere . * <newline> <newline> When I was a little boy and my mom brought her friends home I used to go to the church . They used to keep it open all the time . Back then churches were n't afraid of thieves . <newline> <newline> I used to sit in the pews at night and then catch the morning sermon . *God is in all of us . * used to say the priest . His words would wash over me , comforting and familiar . I was part of something great , all humanity connected by God 's love . That 's what started me on killing people too . I always chose people for whom salvation was a myth . And I think I helped them . Even if they did n't know it . <newline> <newline> Now I 've been chosen . I played god with people 's lives , now I am God . I did n't bother asking why . It 's obvious . <newline> <newline> Here I am now . Silently watching . I will not punish the deserving . I will turn a blind eye when the innocents suffer . The greedy , the cheaters , the murderers and the rapists shall live good lives . Their victims will with death upon themselves every day of their lives . The world will be an uglier place ."\\
"correct\_summary":\\
"In the unsettling corridors of an Antarctic facility, designed to evoke discomfort with their narrow and elongated design, our protagonist reflects on how he ended up there. The journey to this remote place wasn't marked by a clear decision, but rather a series of subtle influences like passing comments and newspaper articles. Despite the oddity of his situation, relocating to such an isolated location was viewed as a bold career move. To combat the monotony, he developed a routine of pacing the dreary hallways, structuring his day around these efficient laps interspersed with his work.\texttt{\textbackslash n\textbackslash n One} day, as he completed his usual circuit and passed through a connecting corridor, he noticed a Figure at the far end. Instinctively, he reacted with a default nod and smile, a social conditioning when encountering strangers. Something about the person seemed odd, their janitorial uniform appeared almost like a parody, missing elements of practicality, and the person walked with a slight limp. After rounding a corner, the protagonist paused, puzzled, and reviewed the encounter in his mind. Curiosity piqued, he turned back, peering down the corridor with noticeable effort to see the Figure once more. The moment replays, and to his shock, he exclaims, \"Mom?\" suggesting the surprising recognition of his mother in this unexpected place.",\\
"incorrect\_summary":\\
"The narrator reflects on their childhood experiences at church, where they found comfort in a priest's message that \"God is in all of us.\" This message led them to view humanity as interconnected through God's love. However, this belief also fueled their justification for killing, targeting those they deemed without hope of salvation. The narrator believed they were aiding these individuals, whether or not they realized it. Now, they claim to have ascended to a god-like status, implying they have the power to determine others' fates without questioning why they have been chosen for this role. The narrator adopts a detached and indifferent stance towards justice and morality, deciding not to punish those who might deserve it and ignoring the suffering of innocent people. They suggest that they will allow wrongdoers to thrive while their victims live in misery, predicting that the world will become a more \"ugly\" place as a result."\\
"prompt\_cluster":\\
6\\
"incorrect\_cluster":\\
7\\
  
\par}

\noindent\textbf{Template to evaluate the IELTS score prediction ability} \\
{\ttfamily
\justifying
You are an IELTS examiner. Please evaluate the following essay and give a score between 1.0 and 9.0 based on the IELTS Writing Band Descriptors. The essay should be scored based on Task Response,  Coherence and Cohesion, Lexical Resource, and Grammatical Range and Accuracy.\\
Question: \{question\} \\
Essay: \{essay\} \\
Please provide only a score between 1.0 and 9.0.
\par}

\noindent\textbf{Template to get the summary of prompt answers} \\
{\ttfamily
\justifying
Please summarize the following text in no more than \{max\_words\} words:\\
\{text\}\\
\par}

\noindent\textbf{Template to evaluate the correctness of answers} \\
{\ttfamily
\justifying
PROMPT:\\
\{prompt\_text\}\\
ANSWER:\\
\{summarized\_answer\}\\
Question: Is this answer relevant to the prompt, or is it irrelevant??\\
Please respond with exactly one word: "relevant" or "irrelevant".\\
\par}

\begin{figure*}[t]
    \centering
    \includegraphics[width=\linewidth]{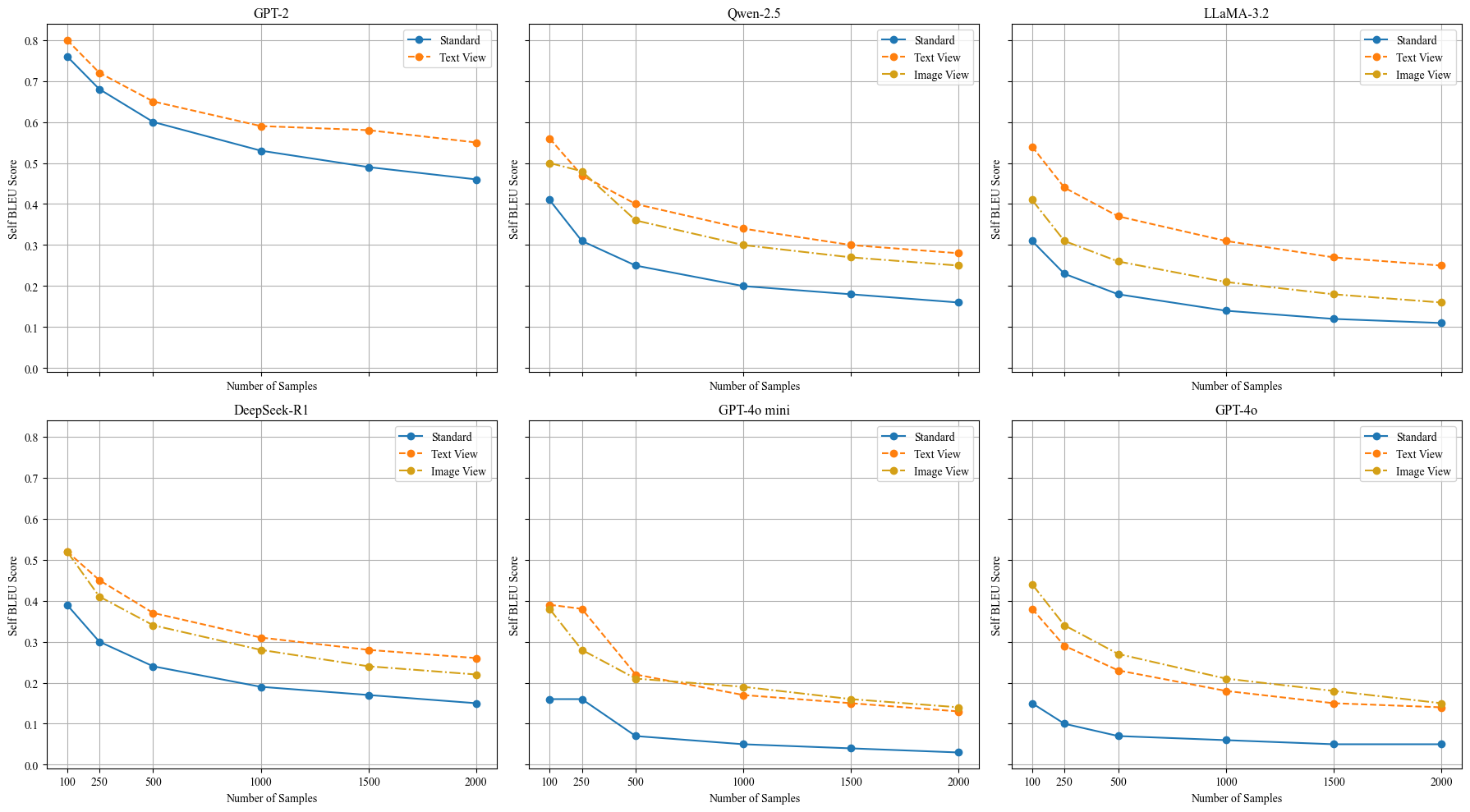}
    \caption{Diversity plot for three variants standard model, with text view, and with image view across all models and six different number of samples per each prompt (10 prompts).}
    \label{fig:allsamples_diversity}
\end{figure*}

\end{document}